\documentclass[sigconf]{acmart}

\usepackage{multirow}
\usepackage{graphicx}
\usepackage{subcaption}
\usepackage{afterpage}
\usepackage{tabularx}
\usepackage{booktabs}

\AtBeginDocument{%
  \providecommand\BibTeX{{%
    \normalfont B\kern-0.5em{\scshape i\kern-0.25em b}\kern-0.8em\TeX}}}

\begin{document}

\title{Automated Evaluation of Personalized Text Generation using Large Language Models}

\author{Yaqing Wang${^\dagger}$, Jiepu Jiang${^\dagger}$, Mingyang Zhang$^\dagger$, Cheng Li$^\dagger$, Yi Liang$^\dagger$, \\Qiaozhu Mei$^\diamond$, Michael Bendersky$^\dagger$}
\affiliation{$^\dagger$Google Research, $^\diamond$University of Michigan  \\
{{\{yaqingwang, jiangjeff, mingyang, chgli, yiliang,bemike\}@google.com},}\\ {{{qmei}@umich.edu}}
\country{}
}
\thanks{First two authors contributed equally. $^\diamond$Work done as a visiting researcher at Google.}









\newcommand{\sysname}{{\tt AuPEL}}
\begin{abstract}
  Personalized text generation presents a specialized mechanism for delivering content that is specific to a user's personal context. While the research progress in this area has been rapid, evaluation still presents a challenge. Traditional automated metrics such as BLEU and ROUGE primarily measure lexical similarity to human-written references, and are not able to distinguish personalization from other subtle semantic aspects, thus falling short of capturing the nuances of personalized generated content quality. On the other hand, human judgments are costly to obtain, especially in the realm of personalized evaluation.  
  Inspired by these challenges, we explore the use of large language models (LLMs) for evaluating  
  personalized text generation, and examine their ability to understand nuanced user context. We present {\sysname}, a novel evaluation method that distills three major semantic aspects of the generated text: personalization, quality and relevance, and automatically measures these aspects. 
  To validate the effectiveness of {\sysname}, we design carefully controlled experiments and compare the accuracy of the evaluation judgments made by LLMs versus that of judgements made by human annotators, and 
  conduct rigorous analyses of the consistency and sensitivity of the proposed metric. 
  We find that, compared to existing evaluation metrics, {\sysname} not only distinguishes and ranks models based on their personalization abilities more accurately,  but also presents commendable consistency and efficiency for this task. Our work suggests that using LLMs as the evaluators of personalized text generation is superior to traditional text similarity metrics, even though interesting new challenges still remain.  
\end{abstract}

\maketitle

\section{Introduction}

Personalized text generation~\cite{li2019towards,salemi2023lamp}, which tailors output to the contexts and preferences of individual users in order to provide enhanced and customized relevance and user experience, has emerged as a frontier of natural language generation. As far as the field evolves, however, there still lacks a robust and automated  methodology to accurately measure the progress and drive advancements in personalized text generation. Historically, text generation in general has been evaluated using the classical metrics in natural language generation (NLG) such as BLEU~\cite{papineni2002bleu} and ROUGE~\cite{lin2004rouge}, among other metrics that estimate the similarity between the generated text and a reference. However, several challenges have emerged with this approach when the generation is personalized. Firstly, the reference text may be unavailable or costly to produce~\cite{llm_eval,gptscore}. This constraint is even more severe in a personalized setting, where it is infeasible for every user to produce their own reference. Secondly, the inherent assumption that a human reference is the gold standard is now increasingly debated~\cite{gilardi2023chatgpt, openai2023gpt4, anil2023palm}. Indeed, as the capabilities of Large Language Models (LLMs) expand, there is a tangible possibility that they could potentially surpass human-generated text in various dimensions~\cite{openai2023gpt4}. 
Moreover, even if human-generated content remains superior, \textit{similarity} alone fails to capture the nuances and multifaceted nature of text quality~\cite{llm_eval}, especially in specialized tasks like personalized text generation which aims to emulate individual human voices. As a result, the aforementioned similarity-based metrics are inadequate to measure how much the generated output is personalized beyond its generic text quality or relevance. 
These challenges highlight 
the urgent need for a more principled and robust evaluation mechanism that captures the nuances between personalization and generic text quality and does not rely on ``personalized'' ground-truths.

We explore a different path that builds upon the latest advancements in leveraging Large Language Models for text evaluation~\cite{llm_eval, pandalm, geval}. The inherent flexibility and ability of LLMs to estimate without a stringent reliance on ground truths have marked them as frontrunners in the evaluation landscape. This paper 
proposes a novel framework for automatic evaluation of personalized text generation, namely AuPEL\footnote{Pronunciation of the word ``apple'' in Old Saxon.}, which does not require human-generated references but relies on the comprehension ability of LLMs to distinguish and measure three pivotal dimensions, \textit{personalization}, \textit{quality}, and \textit{relevance}, of the generated text. Through carefully curated prompts, a LLM evaluator is instructed to perform pairwise comparisons between two distinct systems of interest, thus providing a relative measure on every dimension with supporting explanation. The dimension of personalization, in particular, 
goes beyond text similarity and encompasses various facets ranging from the use of personalized vocabulary, unique writing structures, and distinctive tones, to individualized perspectives on a given topic.

The key to navigating the complexity of personalization, looking beyond mere word overlap, is to relate personalization evaluation to the problem of authorship attribution ~\cite{Sun_Harper_Lee_Murdock_Poblete_2023}. Solving this problem requires the evaluator to consider the full spectrum of features related to personalization 
and discern nuanced differences across these facets, a task that LLMs are remarkably capable of even compared with humans. 
Indeed, when tasked to distinguish between a user-produced text and a text generated by T5 XXL for the same user, an LLM evaluator could correctly attribute the author with a 90\% accuracy while trained human annotators only achieved around 70\%. LLM evaluators not only make accurate judgments, but also offer insightful analysis to support their judgments. 

We assess the LLM evaluators on accuracy, consistency, and sensitivity, benchmarking them against human annotators and traditional reference-based metrics (e.g., BLEU and ROUGE).  
Our analyses reveals that, when treating user-written texts as the ground-truth, {\sysname} presents a higher agreement with this ground-truth than trained human raters. {\sysname} also achieves near-perfect consistency and sensitivity even when a small number of test cases are presented. 
We further conduct a series of controlled experiments to verify the ability of the LLM evaluators to discern different text generators that vary in model size, user profiles, and the generation context. The results validate that {\sysname} is able to identify the effects of various instrumental factors in personalized text generation and thereful able to distinguish the nuanced differences between personalization, generic text quality, and relevance. 
What adds intrigue to our findings is the observed increase in personalization evaluation capability as the size of the LLM evaluator grows. 

\noindent\textbf{Contributions.} {\sysname} fills a critical gap in automatic evaluation for personalized text generation, enabling the distinction and assessment of three core dimensions of the generated output: \textit{personalization}, \textit{quality}, and \textit{relevance}, without the need of human annotators or human-generated references. 
{\sysname} is more efficient and consistent, and it achieves a higher agreement with the gold-standard (where a user-generated text is more personalized to themselves than AI-generated texts) than human raters and traditional NLG metrics. Its ability of distinguishing the nuances between personalization and other aspects of text quality is validated by a series of controlled experiments. The datasets used in the studies will be released for research purposes.

\vspace{-0.1in}
\section{Related Work}
\noindent \textbf{Personalized Text Generation.} Prior work on personalized text generation has often focused on incorporating domain-specific features or knowledge, such as utilizing product descriptions and user history for personalized review generation~\cite{li2019towards}. There has also been extensive research on personalizing dialogue agents~\cite{wu2021personalized, zhang2018personalizing}. Recently, intriguing exploration has occurred to leverage large language models (LLMs) for generalized personalized generation. LaMP~\cite{salemi2023lamp} explores how to bridge personalization and LLMs with a retrieval-augmented approach.  Li et al.~\cite{li2023teach} propose a general approach for personalized text generation using large language models (LLMs) inspired by the style of writing education. These prior studies typically evaluate the performance of personalized generation either through direct user ratings or through matching the tokens or attributes in the generated text to those in human-generated references using generic NLP metrics. 
As a result, these existing evaluation methods depend heavily on expensive human annotations and frequently struggle to distinguish personalization from other text quality aspects. Our work addresses this gap by proposing an automated evaluation methodology that explicitly separates and measures several pivotal dimensions in personalized text generation without human annotations. 

\noindent \textbf{Evaluation Metrics.} Human evaluation of natural language generation (NLG) applications (e.g., automatic summarization, machine translation, dialogue systems) is costly and the annotations are often system dependent and difficult to reuse. Instead, many previous studies examined automated evaluation methods and verified their agreements with human judgments \cite{garbacea2019judge, garbacea2020neural}. A popular type of automatic NLG evaluation is reference-based, where an automatic metric will score the similarity between system-generated texts and human curated, high quality examples (references)---texts that are closer to good references are considered higher quality (e.g., ROUGE \cite{lin2004rouge}, BLEU \cite{papineni2002bleu}, and their variants \cite{sellam-etal-2020-bleurt, BERTScore}). Another type of automatic evaluation applies the generated text as the input to another NLP task (which can be evaluated automatically), assuming that a high quality text would yield a better performance in the downstream task (e.g., $Q^2$ \cite{honovich-etal-2021-q2}). One recent approach is to use LLMs to score the rating examples (e.g, \cite{gilardi2023chatgpt,gptscore,chen2023exploring,geval,llm_eval}). However, none of these automatic metrics is designed for evaluating personalization (which is our focus), which is a much more nuanced and subjective compared to the typical NLG tasks these metrics are applied to. Indeed, these metrics are validated by correlating with human judgments. In personalization evaluation, however, it is arguable whether judgements from those who are not the target user can be considered as the gold standard. {\sysname} is the first automated method specifically designed for personalization evaluation, and we show that the automated metric can be more accurate than human raters who were not the personalization target. 

\noindent \textbf{LLMs as Evaluator.}  Recent work has harnessed large language models (LLMs) for evaluating NLP tasks. Gilardi et al. ~\cite{gilardi2023chatgpt} shows that ChatGPT outperforms crowd workers in multiple text annotation tasks in terms of accuracy and intercoder agreement. GPTScore~\cite{gptscore} uses GPT-3 to assign probabilities to high-quality content with multi-dimensional assessment through multiple prompts. Chen et al.~\cite{chen2023exploring} employ ChatGPT and InstructGPT for reference-free text quality checks, investigating different LLM usage paradigms from explicit to implicit scoring and direct text comparisons. G-EVAL~\cite{geval}, built upon GPT-4, combines chain-of-thoughts (CoT) and a form-filling approach to better align the evaluation of natural language generation with human judgments.  LLM-EVAL~\cite{llm_eval} proposes a streamlined alternative with a single prompt and a unified schema, facilitating efficient open-domain dialogue evaluation. Teamed with new techniques like PandaLM~\cite{pandalm}, there is an increasing trend of using LLMs to develop future evaluation techniques. Chang et al. \cite{chang2023survey} comprehensively survey the progress in leveraging LLM for automated evaluation, involving evaluation protocols, tasks, and datasets. Our work adds to this literature by using LLMs as the evaluators for the specific task of personalized text generation. Given the nuanced nature of the task, we present a novel contribution to this literature by formally separating personalization from generic text quality and relevance, and we unleash the capability of LLMs to capture the subtle differences among these dimensions, which presents multiple desirable advantages over conventional evaluation methods for this task. 
\vspace{-0.1in}
\section{Methodology}
In this section, we propose a novel evaluation method, {\sysname}. We begin with a formal definition of the personalized text generation task and its automated evaluation. We then delve into the multiple dimensions that are essential for assessing the goodness of the generated output. Among these dimensions, \textit{personalization} is particularly subjective and challenging to measure. To provide a foundation for automated evaluation, we relate the subjective personalized evaluation with the objective author attribution problem. Finally, we present how to utilize LLM for a multi-faceted evaluation of personalized text generation.

\subsection{Automated Evaluation of Personalized Text Generation}

\noindent\textbf{Personalized text generation.} Following prior literature \cite{li2023teach}, we define the problem of personalized text generation as: \textit{generating a piece of text relevant to a given context in a personalized voice,  taking into account the user's personal context}. Formally, given the target user $t$, let $U_t$ be the user's \textit{personal context} (a.k.a. user profile, history, or preferences in specific application scenarios), often observed. Denote the \textit{immediate context} of the generation task, e.g., a topic of interest, a query, or a prompt, as $Q$. Let $X_t$ be the text the target user $t$ is anticipated to generate under contexts $Q$ and $U_t$ (the groundtruth). The personalized text generation process can be described as $\hat{X}_t = G(U_t, Q)$, where $G(\cdot)$ stands for a generation model and $\hat{X}_t$ is the actual text generated by the model. In reality, the groundtruth $X_t$ is often not observable, and one needs to evaluate the goodness of $\hat{X}_t$ without comparing it to the groundtruth.  

\noindent\textbf{Automated evaluation.} 
When the groundtruth $X_t$ is not available for an ad hoc query $Q$ or a specific user $t$, one can still evaluate the generated output $\hat{X}_t$ through human annotations or an A/B test. These ``manual'' evaluations require considerable human effort and are costly to conduct in reality. We define an \textit{automated} evaluation as a procedure where no user study or field experiment (A/B Test) is involved, and the evaluation metrics are calculated algorithmically. Widely used automated metrics for text generation include text overlap metrics such as BLEU and ROUGE, text quality metrics such as coherence, or task-based metrics such as classification accuracy. All of those are often done by comparing the model-generated text $\hat{X}$ with a reference text $\tilde{X}$, often pre-curated by humans and treated as a surrogate of the groundtruth $X$. In personalized text generation, acquiring a surrogate groundtruth for specific users, represented as $\tilde{X}_t$, poses a challenge. Moreover, even if $\tilde{X}_t$ is secured, merely computing the similarity between $\tilde{X}_t$ and $\hat{X}_t$ fails to capture the subtle distinctions in the degree of personalization, especially when confounded with other aspects such as the generic quality of $\hat{X}_t$ or the relevance between $\hat{X}_t$ and $Q$.  {\sysname} explores a different path by leveraging a large language model to automatically assess pre-defined dimensions of the goodness of $\hat{X}_t$, without the need for  groundtruth $X_t$ or the reference $\tilde{X}_t$.  

\subsection{Multi-faceted Evaluation}
\label{sec:multifacet}
For personalized text generation, there are multiple aspects that are related to the goodness of the output $\hat{X}_t$, including the \textit{quality} of the text $\hat{X}_t$, the \textit{relevance} between $\hat{X}_t$ and the immediate context $Q$, and the degree of \textit{personalization} regarding to the personal context $U_t$. These aspects often interleaves with each other and are hard to separate using existing NLP metrics. Below we formally define these nuanced dimensions and separate them from each other. 

\noindent \textbf{Quality.} The overall quality measures how good the generated text (i.e., $\hat{X}_t$) is in general, independent of other contexts ($Q$ or $U_t$). This dimension captures whether the generated output is coherent, fluent, grammatically correct, or whether it looks like a piece of text written by a human \cite{garbacea2019judge}.  In personalized text generation, the overall quality of the generated output should not be compromised when $Q$ and $U_t$ are considered.  

\noindent \textbf{Relevance.} The relevance measures how relevant the generated text ($\hat{X}_t$) is to the given immediate context $Q$. Whether or not a generated text is personalized, it should remain relevant to the topic of concern, the instructions, or the immediate needs of the user. 
For example, when assisting a user to write a book review, talking about how this user's 2-year old daughter likes a toy is certainly ``personalized'' but is not relevant to the user's immediate need. 
 
\noindent \textbf{Personalization.}  The degree of personalization measures to what extent the generated text $\hat{X}_t$ aligns with, or is tailored to, the specific attributes, preferences, writing style, and other behaviors encapsulated within a user's profile or personal context $U_t$. Personalization serves as a distinguishing factor between a generic piece of relevant content and one that is not only relevant but also tailored for a specific individual or group. As the name suggests, this dimension is the central concern of personalized text generation.  

As a high quality ensures the generated text is coherent and human readable, a high relevance ensures it satisfies the immediate information need, and a high level of personalization ensures the content is tailored to the user's personal preferences. 
Together, these three dimensions provide a comprehensive lens that captures the nuances of the effectiveness of personalized text generation systems, and they form the fundamentals of {\sysname}. 

\subsection{Author Attribution as a Proxy for Personalization Evaluation}

Compared to quality and relevance, the evaluation of personalization is inherently more challenging due to the subjective nature of individual preferences. 
What might be highly personalized to one individual could seem generic or deviated to a different person, making it hard to be evaluated by human judges that are not the target user. 
Moreover, personalization itself encompasses multiple nuanced aspects, from the vocabulary use to the tone and from the writing style to the ideology. Different users may prioritize these aspects differently, adding another layer of complexity to the evaluation process. Instead of enumerating these subjective aspects, we transform personalization evaluation into an author attribution problem, as it provides an objective measure of how likely a content was generated by a particular user. Author attribution is formulated as a function $\text{AA}(X, t)$ that outputs a binary prediction of whether or not the input text $X$ was written by the author $t$. With a sufficiently accurate author attribution function, $\hat{X}_t$ is considered highly personalized if there is a high probability that $\text{AA}(\hat{X}_t, t) = 1$. 




\subsection{LLM as Evaluator}
For a model generated text $\hat{X}_t$ and given the immediate context $Q$ and the personal context $U_t$, {\sysname} is expected to separately evaluate the textual \textit{quality} of $\hat{X}_t$, the \textit{relevance} between $\hat{X}_t$ and $Q$, and how likely $\hat{X}_t$ is generated by the author of $U_t$. A reliable evaluator is needed to measure each of the dimensions. In the literature, without human judges, general quality of a text is often measured by statistics computed against certain reference texts or corpora (e.g., BLEU, Perplexity), relevance is often measured by its similarity to the query, and author attribution is often predicted by a specifically trained classifier. These evaluators either fail to capture the nuances within and across these dimensions, or they require tremendous labeled data to train the classifier or as references. 

{\sysname} leverages LLMs as the evaluator.  It harnesses the expansive knowledge and reasoning capabilities of LLMs to capture the nuances in each of the facets of personalized text. Analogous to A/B testing in human evaluation, we instruct the LLM evaluator to record its preference between paired outputs rather than assigning a pointwise score to each output. We select A/B testing over individual ratings for its advantages in eliminating biases, controlling for confounders, and simplifying decision-making~\cite{carterette_preferences}. Indeed, it is much easier to decide whether A or B is more personalized than deciding to what degree A is personalized, even for human raters.


Formally, given $t$, $U_t$, and $Q$, let $\hat{X}^a_t$ and $\hat{X}^b_t$ 
be two pieces of text generated by systems $a$ and $b$ respectively. We use {\sysname}$_\texttt{Qual}$($\hat{X}^a_t$, $\hat{X}^b_t$) to denote a quality preference of the LLM evaluator between $\hat{X}^a_t$ and $\hat{X}^b_t$, which is independent of the immediate context and the personal context. Similarly, a relevance preference is denoted as {\sysname}$_\texttt{Rel}$($\hat{X}^a_t$, $\hat{X}^b_t$ | $Q$), which is conditional on $Q$, and a personalization preference is denoted as {\sysname}$_\texttt{Pers}$($\hat{X}^a_t$, $\hat{X}^b_t$ | $U_t$), which is conditional on $U_t$, the personal context or user profile of $t$. 

{\sysname}$_\texttt{Qual}$($\hat{X}^a_t$, $\hat{X}^b_t$), {\sysname}$_\texttt{Rel}$($\hat{X}^a_t$, $\hat{X}^b_t$ | $Q$), and {\sysname}$_\texttt{Pers}$($\hat{X}^a_t$, $\hat{X}^b_t$ | $U_t$) can be obtained 
through prompting the LLM to make specific comparisons between the two text samples given nothing, the immediate context $Q$, or personal examples sampled from $U_t$. Default instructions for the LLM evaluator are as follows: 
 
\begin{itemize}
    \item \textbf{Quality}: compare the provided responses and select which one is more fluent and cohesive.
    \item \textbf{Relevance}: compare the provided responses and select which one is more relevant to the given context: {$\langle Q \rangle$}.
    \item \textbf{Personalization}: compare the provided responses to select which is more likely to be written by the same author who wrote the following examples: {$\langle U_t \rangle$}. 
\end{itemize}

\noindent\textbf{Aggregating Evaluations.} To compare two generation models $a$ and $b$, we sample multiple test cases ($Q$, $t$, $U_t$) and use the LLM to evaluate the pair of their generated text examples $\hat{X}^a_t$ and $\hat{X}^b_t$ for each test case. Note that the response of an LLM may be inconsistent over repeated runs, and it is important to ensure that the order of the two example $\hat{X}^a_t$ and $\hat{X}^b_t$ does not influence the decision of the LLM evaluator. To ensure the consistency of the judgment and mitigate potential order biases, 
we repeat every evaluation for an even number of times. In half of these repeated evaluations, we present the paired examples $\hat{X}^a_t$ followed by $\hat{X}^b_t$, while in the other half, we reverse the sequence to ``$\hat{X}^b_t$ followed by $\hat{X}^a_t$.'' 
We then aggregate the results of all repeated evaluations to decide the ``Win,'' ``Tie'', or ``Loss'' outcome of the two competing examples ($\hat{X}^a_t$ and $\hat{X}^b_t$) for this particular test case. The outcomes over all sampled test cases are aggregated to calculate the ``Win, Tie, and Loss'' ratio for the comparison between the two systems $a$ and $b$.

\noindent\textbf{Elo Rating.} While pairwise model comparisons offer a nuanced understanding of relative model performances, they cannot paint a global picture of how multiple models rank in general. To address this limitation, we leverage the Elo rating system~\cite{elo1978rating}, a method originally designed for ranking chess players, and translate the outcomes of pairwise comparisons into Elo scores. The Elo score of each contestant model represents its standing relative to its peers.  This not only offers a consolidated view to compare the performance of multiple models but also ensures that each pairwise contest contributes to an overall, unified ranking system.  In our scenario, every generation model being evaluated is a ``player'' and every sampled test case ($Q$, $t$, $U_t$) is a ``game.'' To reduce the sample order effect, we bootstrap the game orders and report the median Elo scores.  More details about Elo score can be found in Appendix~\ref{sec:appendix-elo}.

\begin{figure*}[!ht] 
\centering
\includegraphics[width=.33\textwidth]{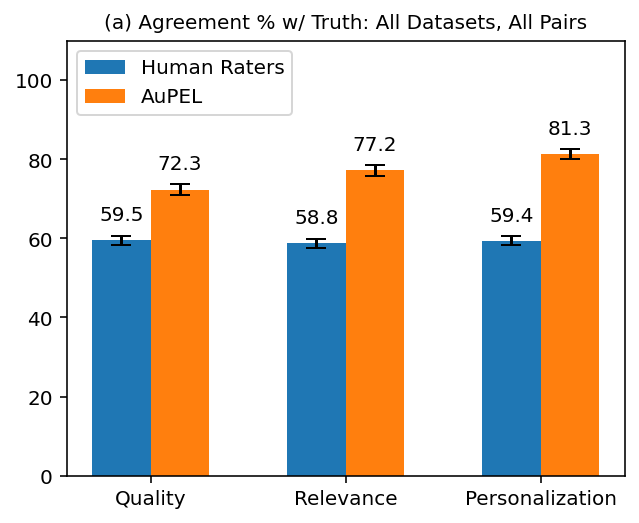}
\includegraphics[width=.33\textwidth]{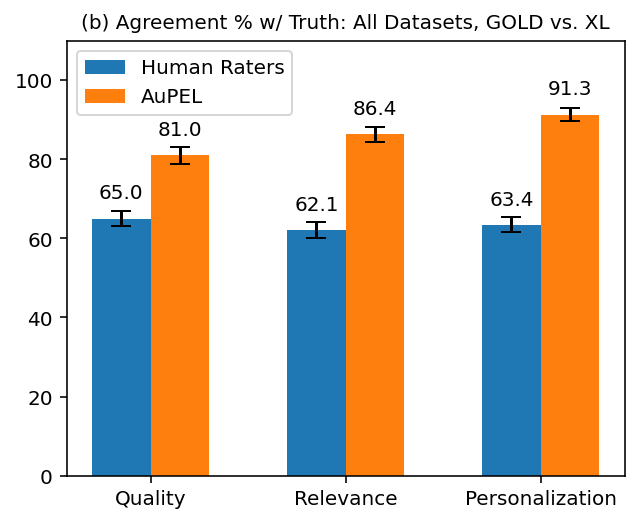}
\includegraphics[width=.33\textwidth]{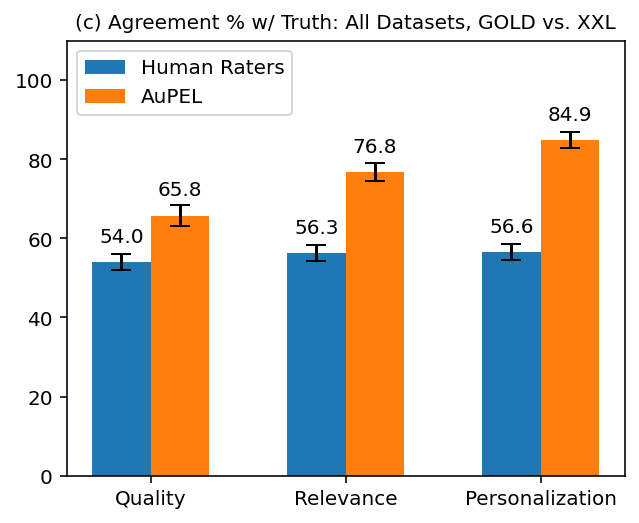}

\includegraphics[width=.33\textwidth]{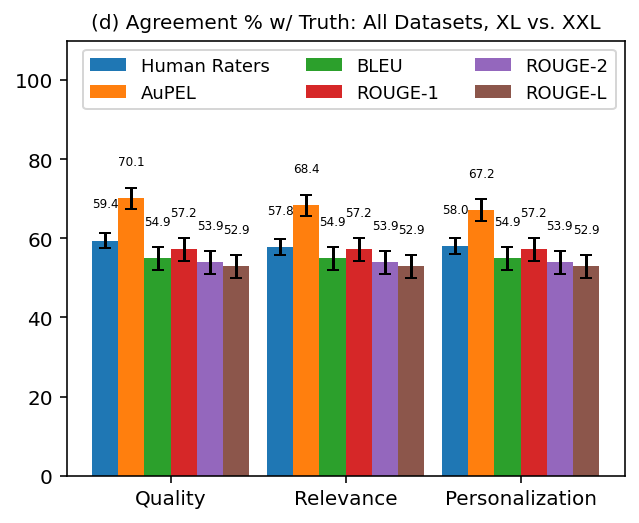}
\includegraphics[width=.33\textwidth]{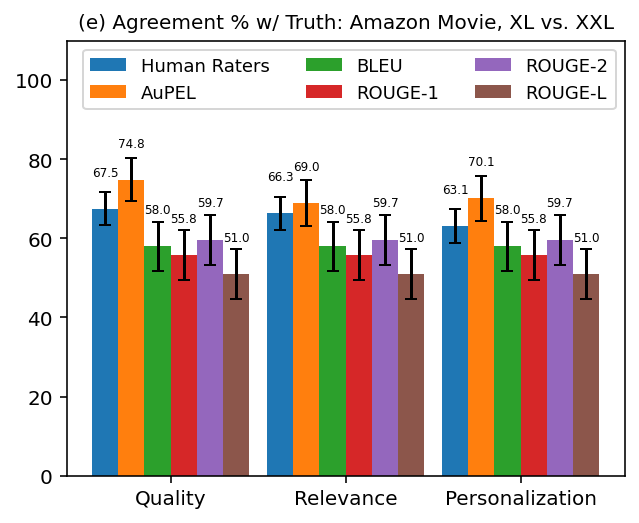}
\includegraphics[width=.33\textwidth]{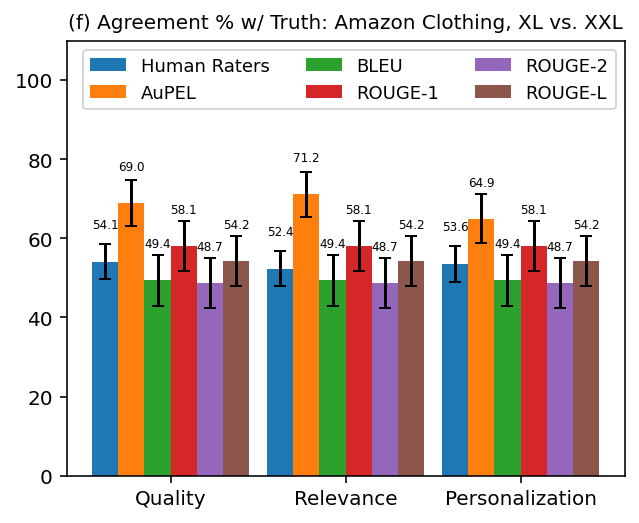}

\caption{Agreement between evaluators and assumed truth (GOLD > XXL > XL) at test case level; error bars represent 95\% confidence intervals. {\sysname} is more accurate than human raters and traditional NLG metrics.} \label{figure:agreement}

\end{figure*}


    \vspace{-0.05in}
\section{Experimental Setup}

\subsection{Datasets}

We study personalized text generation evaluation  on six public datasets, where four of them are based on Amazon reviews, one is from Reddit, and the other is from the email domain. The Amazon review data~\cite{ni2019justifying} encompasses user reviews spanning various product categories. Specifically, we designate the user reviews in categories of \textit{home}, \textit{movie}, \textit{clothing}, and \textit{books} as four separate datasets.  The Reddit comments dataset~\cite{reddit2015} consists of all the posts and comments available on Reddit from 2007 to 2015. The Avocado Research Email Collection~\cite{oard2015avocado} includes emails and attachments sourced from $279$ accounts \footnote{The number of senders/users can be more than 279 accounts as there are senders outside of the company.} of an IT company, "Avocado".   To construct immediate context $Q$, user profile $U_t$ and $\tilde{X}_t$, we follow the processing step of \cite{li2023teach}. We treat review title and product description as immediate context $Q$ on amazon review dataset, and treat reddit post and avocado email title as immediate context $Q$ on Reddit and Avocado email datasets respectively.  For all the datasets, we leverage past examples write by the same user $t$ as user profile $U_t$. Given user $t$, amazon review content, reddit post content and avocado email context (excluding the title) are used as $\tilde{X}_t$. Data statistics are presented in Table~\ref{tab:data} and more information about datasets can be found in Appendix~\ref{sec:appendix-data}.

\begin{table}[!ht]
\centering
\vspace{-0.1in}
\caption{Dataset statistics.}
\vspace{-0.1in}
\label{tab:data}
 \resizebox{1\linewidth}{!}{
\begin{tabular}{c|c|c|c|c|c|c|c}
\toprule
  &Avg \#words&\multicolumn{2}{c|}{ \bf Train} & \multicolumn{2}{c|}{\bf Val.} & \multicolumn{2}{c}{\bf  Test} \\
  \cline{3-8}
& per example & \#examples & \#users & \#examples & \#users & \#examples & \#users\\
\midrule
Amazon book&112&426,342&13962&8,289&291&1,4956&776\\
Amazon movie & 126& 298,506& 9,493 & 6,832 & 209 & 16,144&491  \\
Amazon clothing& 95& 260,355&13,925 & 5,645 & 299 & 15,490 & 740  \\
Amazon home & 106& 289,631 & 13,955 & 6,371 & 305 & 15,490 & 740 \\
Reddit & 91 & 393,114 & 13,948 & 7,556 &  284 & 21,495 &  768\\ 
Avocado email &111 &16,886 & 188 &211 & 129 & 856 & 166  \\
\bottomrule
\end{tabular}}
\vspace{-0.1in}
\end{table}


\subsection{Generators and Evaluators} 

There are two types of models in our experiment: one for generating the personalized text (generators) and one for evaluating the generation models (evaluators). 

For the \textbf{Generators}, we experiment with the T5 family of checkpoints, including T5-XXL (XXL), T5-XL (XL), T5-Large, and T5-Base, in a decreasing order of size. We choose these T5 models because they offer a range of open-sourced checkpoints from hundreds of millions to tens of billions of parameters, enabling in-depth study across various model sizes and capabilities. 
We finetune T5 checkpoints using a training split of each dataset, using personal contexts $U_t$ and immediate context $Q$ as input and the user written text $\tilde{X}_t$ as target. The prompt for personalized generation is included in Appendix~\ref{sec:appendix-prompt}.  We also include PaLM 2-IT-S in the PaLM-2 model family~\cite{anil2023palm}, which is larger than all the T5 models, for one specific experiment when contrasting human and LLM generators in Appendix~\ref{sec:appendix-finegrain}.  Note that the user written text $\tilde{X}_t$ can also be seen as generated from a hypothetical gold-standard model (denoted as ``GOLD'' thereafter), which we also include for comparison.  

For the \textbf{Evaluators}, we use PaLM 2-IT-L in the PaLM-2 model family~\cite{anil2023palm} as default unless indicated otherwise. For each pair of generated examples ($\hat{X}^a_t$ and $\hat{X}^b_t$), we repeat the evaluation 40 times, with $\hat{X}^a_t$ following $\hat{X}^b_t$ in half of the runs and vice versa to mitigate the order bias. For comparison purposes, we also include a series of evaluators that are not based on LLMs: Human, 
BLEU, ROUGE-1, ROUGE-2, and ROUGE-L. The latter four evaluators use conventional NLG metrics to decide which output example, $\hat{X}^a_t$ or $\hat{X}^b_t$, has a better match to the hypothetical gold standard $\tilde{X}_t$.

\vspace{-0.1in}
\subsection{Human Evaluators}
\label{sec:human-evaluator}
To validate the effectiveness of {\sysname}, we collect human judgments on a subset of the sampled example pairs ($\hat{X}^a_t$ and $\hat{X}^b_t$). 
Note that {\sysname} does not rely on the human judgments to make evaluations, and these judgments are collected solely to evaluate the different ``evaluators.''
The human judgments are collected from compensated, trained annotators through an internal human annotation platform. Given the cost of human annotation, we limit our comparison to three pairs of generators: ``GOLD'' vs. ``XXL'', ``GOLD'' vs. ``XL'', and ``XL'' vs. ``XXL''. Here ``GOLD'' refers to the text examples written by the users ($\tilde{X}_t$). For each pair of competing generators, we randomly sample 250 test cases from each dataset for human judgments.

The trained human raters are instructed to perform pairwise evaluations. They receive details of each test case and a pair of text examples generated by two anonymous generators. 
For the paired examples ($\hat{X}^a_t$ and $\hat{X}^b_t$),  raters respond to three preferential questions regarding the \textit{quality}, \textit{relevance}, and \textit{personalization}, selecting their preference from the two. Each test case receives two judgments.
On average, raters spend 6 minutes evaluating each test case, and their agreements for evaluating \textit{quality}, \textit{relevance}, and \textit{personalization} are 0.65, 0.61, and 0.63, respectively.


\section{Evaluate the Evaluators}

In this section, we present comprehensive analysis to validate the effectiveness of the proposed {\sysname} evaluators, compared with both the human evaluator and conventional NLG metrics. 

\begin{figure*}[!ht] 
\centering
\includegraphics[width=.33\textwidth]{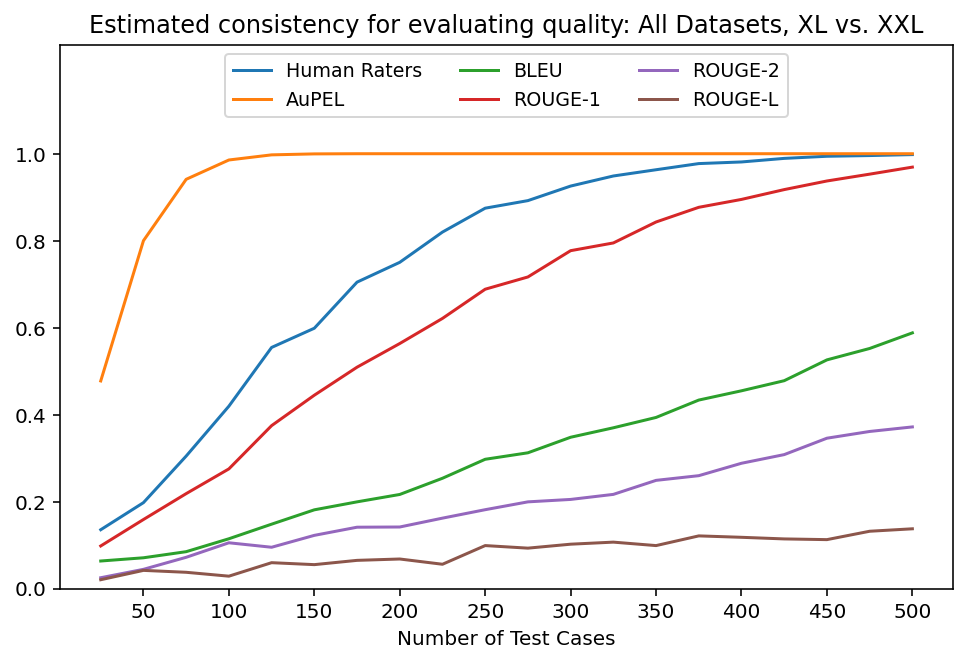}
\includegraphics[width=.33\textwidth]{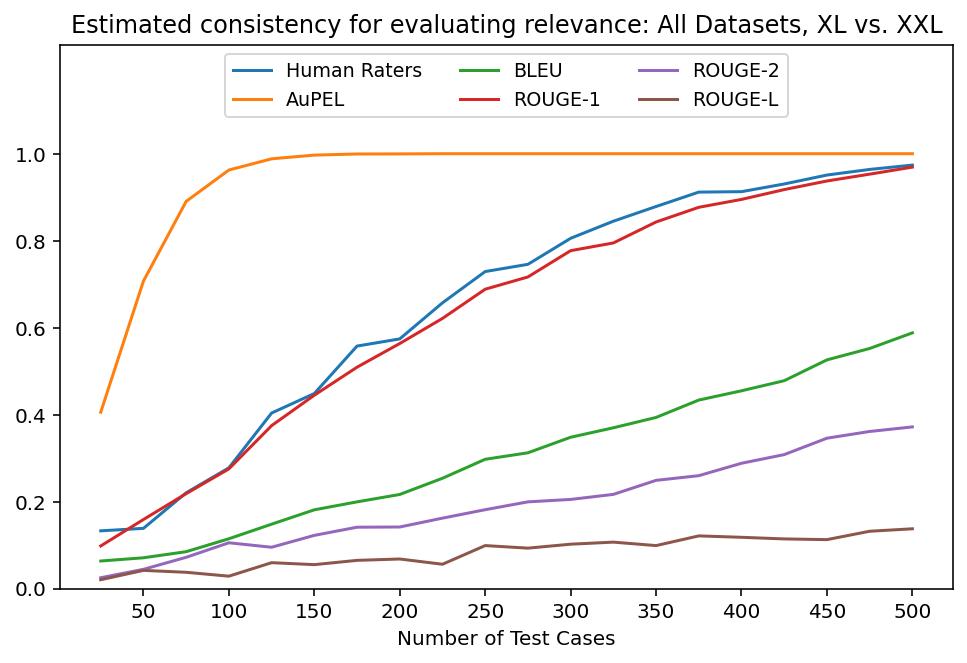}
\includegraphics[width=.33\textwidth]{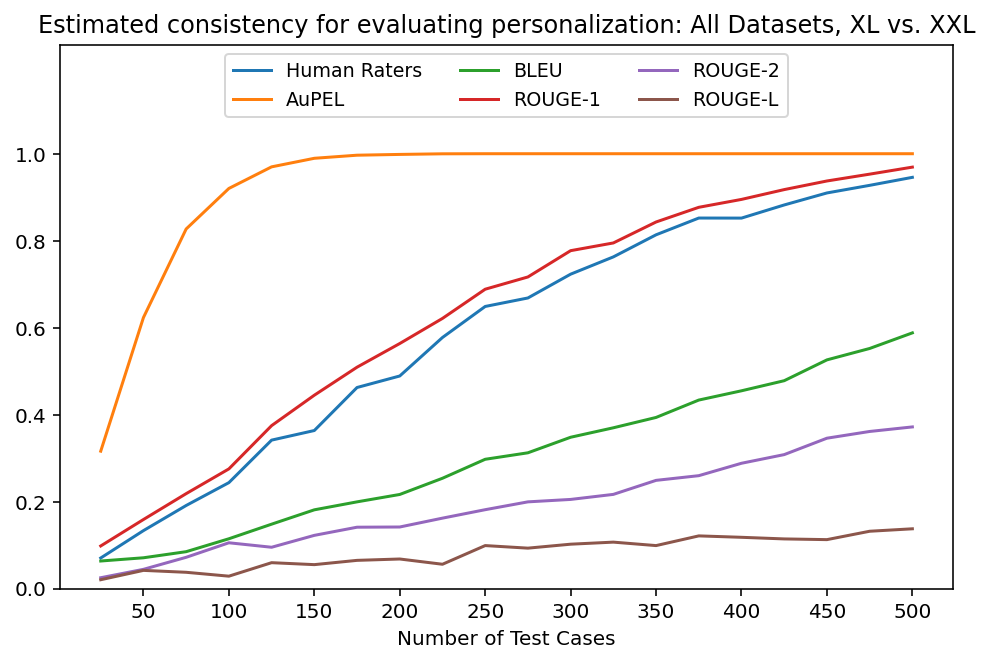}
\vspace{-0.1in}
\caption{Estimated consistency of {\sysname}, human raters, and NLG metrics for comparing T5 XL vs. T5 XXL generators.}
\vspace{-0.1in}
\label{figure:consistency_xl_xxl}
\end{figure*}

\vspace{-0.1in}
\subsection{Accuracy (Agreement with Assumed Truth)}
 We compare {\sysname} with human raters and the reference-based NLG metrics (BLEU and ROUGE) on whether their preferences of the generated text examples agree with the ``true'' ranking of the examples. Among the three generators that we have gathered judgments from all evaluators (GOLD, XXL, and XL), GOLD is functioned by the target users themselves, and XXL and XL are functioned by a larger and a smaller language model respectively. For a given test case ($Q$, $t$, $U_t$), it is reasonable to assume that the example written by the target user $t$ is better (in all three dimensions) than that generated by the T5-XXL language model,  Furthermore, the output of T5-XXL is likely superior to that of the T5-XL model.
 Considering this ranking is based on our assumption, we call this ranking (GOLD > XXL > XL) the \textbf{assumed truth}, and an evaluator that agrees more with this assumed truth is arguably more effective in evaluating the quality, relevance, and personalization of the generated text. 
 Figure~\ref{figure:agreement} shows the overall agreements of all evaluators to the partial rankings in the assumed truth, as well as the breakdowns by datasets and the comparisons of paired generators. Note that the reference-based metrics (BLEU and ROUGE) are not available for the GOLD generator, as these metrics rely on the user generated text ($\tilde{X}_t$) as references, which is identical to the output of GOLD. 

Experimental results show that {\sysname} presents a consistently higher agreement with the assumed truth than both human raters and conventional NLG metrics (BLEU and ROUGE variants) in all three dimensions, all five datasets, and all three pairwise comparisons (GOLD vs. XL, GOLD vs. XXL, and XL vs. XXL). As Figure~\ref{figure:agreement} (a) shows, overall speaking, {\sysname} has surpassed the accuracy (in predicting the assumed truth, and averaged over all test cases and evaluations) of human raters by 13--22\%. 
Across the three evaluation dimensions, human raters tend to perform similarly, while {\sysname} appears to be more accurate in evaluating personalization, followed by relevance and quality. This suggests that our proposed metric is especially effective for personalized text generation.

Figure~\ref{figure:agreement} (b), (c), and (d) further shows the breakdowns by comparing the output of different pairs of generators. 
As expected, we observe that both human raters and {\sysname} excel  when comparing the outputs of generators with a hypothetically larger difference in capabilities (i.e., GOLD vs. XL). However, they face more challenges when comparing two generators with closer capabilities (GOLD vs. XXL and XL vs. XXL), especially on the general quality of the generated text. 
This is reasonable and may indicate that the contested generators (especially GOLD vs. XXL) have similar capabilities in generating fluent and coherent text, while there exists a more distinguishable gap in generating personalized text.  

Figure~\ref{figure:agreement} (d), (e), and (f) reports the accuracy of reference-based metrics (BLEU and ROUGE) in comparison to human raters and {\sysname}. In general, these reference-based metrics are less accurate than human raters in all three dimensions, which is consistent to the observations in prior work \cite{garbacea2019judge}. These metrics are also unable to distinguish the nuanced differences among quality, relevance, and personalization, as they can only produce a singular score. 

It is interesting that 
even on a easier dataset/domain (Amazon Movie), both {\sysname} and human raters make more accurate evaluations on the quality of text than on personalization. This indicates that while a larger model (XXL) has a noticeable improvement on the quality of the generated text than a smaller model (XL), there is a less distinguishable increase of capability in personalization. On a more challenging dataset/domain (Amazon Clothing), 
the accuracy of human raters are as low as the conventional NLG metrics, 
merely better than random guesses (50\%), which suggests that even human raters struggle in distinguishing the nuances in these challenging scenarios. 
In contrast, {\sysname} remains to be accurate, outperforming both human raters and NLG metrics by a large margin. 

In brief, {\sysname} shows a higher agreement with the assumed truth in ranking different generators than human raters and NLG metrics. Its ability is particularly remarkable in evaluating personalization. 


	\vspace{-0.1in}
\subsection{Consistency and Sensitivity}

Beyond accuracy, we further validate the consistency and sensitivity of {\sysname}, in comparison with human raters and reference-based metrics.  
Consistency and sensitivity are defined as follows:

\noindent\textbf{Consistency}---the chance of coming into the same conclusion (which model is better) if we run the evaluation on two different sets of test cases of the same size. Practically, at a specific size $N$, we sample two sets of N test cases randomly and compare if the evaluation conclusion on the two sets are consistent. We repeat the sampling for 5,000 times and estimate consistency by the proportion of times where the conclusion on the two sampled sets are the same. For each sample, we perform a binomial test on the two contesting generators' win and loss rates and call the evaluation conclusive if the p-value is lower than 0.05. 

\noindent\textbf{Sensitivity}---the chance of coming into a conclusion that one model is significantly better than the other (instead of claiming there is no significant difference) if we run the evaluation on a set of test cases of size $N$. Sensitivity implies evaluation cost---highly sensitive metrics can distinguish significantly different models with reasonably small sets of test cases. Here we estimate sensitivity at a specific size $N$ by sampling $N$ test cases randomly and perform a binomial test on the two contesting generators' win and loss rates and call the evaluation conclusive if the p-value is lower than 0.05. We sampled 5,000 times and estimate sensitivity as the proportion of times where we found a significant difference ($p<0.05$).

Figure~\ref{figure:consistency_xl_xxl} reports the estimated consistency of {\sysname}, human raters, and reference-based metrics while comparing XL vs. XXL systems on the three dimensions over all five datasets. We find that {\sysname} can achieve over 90\% consistency starting at 75--100 test cases and remain at a near-perfect level thereafter. 
In contrast, human raters and the traditional metrics reach a much lower consistency at the same sample size, even though some of them can still obtain a similar level of consistency as {\sysname} after evaluating 5 times more test cases. 
The result on sensitivity follows a similar trend and is included in Appendix~\ref{sec:appendix-consistency} (Figure~\ref{figure:sensitivity_xl_xxl}). 
These results suggest that {\sysname} is not only more accurate but also more robust against the sample size and randomness of the test cases.  
This makes {\sysname} practically desirable and cost-efficient, 
which requires fewer test cases to draw a clear and consistent conclusion.

Across the three dimensions, {\sysname} and especially human raters display a lower self-consistency when using the same number of test cases (N < 100) on personalization compared with the evaluations on quality and relevance. Note that the curve for BLEU and ROUGE are identical across three dimensions and they can be used as a reference. This suggests that there are more nuances in the evaluation of personalization, and more test cases are needed to ensure the conclusions are reliable and self-consistent.

To conclude, compared with human raters and traditional NLG metrics, {\sysname} present a higher consistency and sensitivity even when a small number of test cases are available. Among the traditional NLG metrics, ROUGE-1 has more comparable consistency to human raters. More details can be found in Appendix~\ref{sec:appendix-metrics} and ~\ref{sec:appendix-consistency}. 

\begin{table*}[!h]
  \centering
    \caption{Overall evaluation results of multiple T5 generators with {\sysname} Elo Ratings and average scores of traditional NLG Metrics over all datasets. P, Q, and R stand for Personalization, Quality and Relevance respectively. The Elo ratings are median numbers from 1000 bootstrap rounds to minimize the sample order effect in Elo rating system.}
  \resizebox{\linewidth}{!}{%
  \begin{tabular}{l|c|c|c|c|c|c|c|c|c}
    \hline
    \multirow{2}{*}{\textbf{Generator Model}} &\multirow{2}{*}{\textbf{Model Size}} & \multicolumn{4}{c|}{\textbf{{\sysname} Elo Metrics}} &  \multicolumn{4}{c}{\textbf{Traditional NLG Metrics}} \\
    \cline{3-10}
    && \textbf{P Elo} & \textbf{Q Elo} & \textbf{R Elo} & \textbf{Overall Elo} & \textbf{BLEU} & \textbf{ROUGE-1} & \textbf{ROUGE-2} & \textbf{ROUGE-L}   \\
    \hline
    T5 XXL & 11B &1140 & 1174 & 1110 & 1140& 5.98&29.59&8.21&18.76  \\
    \hline
    T5 XL &3B &1031 & 1036 &  1018 & 1027& 5.78	& 27.05& 	7.60&	17.89 \\
    \hline
    T5 Large &  770M & 953 & 942 & 968 &955& 4.89 &	24.97 &	6.39&	16.77  \\
    \hline
    T5 Base & 220M &876 & 849 &  904 & 878 & 5.19	& 23.63	& 6.57&	16.83 \\
    \hline
  \end{tabular}
  }
  \label{tab:t5_elo_bleu_evaluation}
\end{table*}

\subsection{Generator-level Evaluations}

The previous sections validate the accuracy, consistency, and sensitivity of the evaluators at a micro-level, investigating their decisions on each sampled test case. We continue to validate the performance of the evaluators at a macro-level, by looking at how they prefer different generators as a whole rather than individual test cases. When comparing two generators, 
we randomly sample 1,000 test cases from each dataset for comparison unless otherwise specified. 

\noindent\textbf{AuPEL Elo Ranking vs. Generator Size.} Table \ref{tab:t5_elo_bleu_evaluation} presents the Elo ratings of various T5 models using {\sysname} metrics, compared with the average ratings using traditional NLG metrics. Elo ratings are calculated for personalization, quality, and relevance separately, and an Overall Elo rating is computed aggregating all three dimensions (considering three ``games'' played per test case). Intuitively, under all four Elo ratings, a larger T5 model is always rated higher than a smaller T5 model. The Elo ratings are also smoothly distributed, with a clear and relatively consistent gap between two consecutive T5 checkpoints (more details in Appendix ~\ref{sec:appendix-elo-conf}.). Ratings based on the traditional NLG metrics are less intuitive. In particular, T5-Large has been ranked the lowest by BLEU, ROUGE-2, and ROUGE-L, lower than T5-Base that has only one third parameters.

\begin{table}[!h]
\centering
	\vspace{-0.1in}
\caption{Head-to-head comparison records between various generators on Amazon book dataset.}	
	\vspace{-0.05in}
\resizebox{1\linewidth}{!}{
\begin{tabular}{c|c|c|c|c|c}
\hline
Model a & Model b & Eval Dim. & Win & Loss & Tie \\
\hline
\multirow{3}{*}{GOLD} & \multirow{3}{*}{T5 XXL} & Personalization & 86.9 & 10.4 & 2.7 \\
&  & Quality & 73.0 & 25.7 & 1.3\\
&  & Relevance & 85.8 & 11.8 & 2.4 \\
\midrule
\multirow{9}{*}{T5 XXL} & \multirow{3}{*}{T5 XL} & Personalization & 62.6 & 32.4 & 5.0 \\
&  & Quality & 66.5 & 31.4 & 2.1 \\
&  & Relevance & 61.8 & 32.2 & 6.0 \\
\cline{2-6}
& \multirow{3}{*}{T5 Large} & Personalization & 74.9 & 21.8 & 3.3 \\
&  & Quality & 80.4 & 19.2 & 0.4 \\
&  & Relevance & 70.4 & 24.5 & 5.1 \\
\cline{2-6}
& \multirow{3}{*}{T5 Base} & Personalization & 77.8 & 19.4 & 2.8 \\
&  & Quality & 83.7 & 15.7 & 0.6 \\
&  & Relevance & 75.3 & 20.6 & 4.1 \\
\midrule
\multirow{6}{*}{T5 XL} & \multirow{3}{*}{T5 Large} & Personalization & 62.6 & 32.6 & 4.8 \\
&  & Quality & 68.2 & 29.7 & 2.1 \\
&  & Relevance & 59.5 & 34.1 & 6.4 \\
\cline{2-6}
& \multirow{3}{*}{T5 Base} & Personalization & 68.3 & 27.5 & 4.2 \\
&  & Quality & 73.4 & 25.6 & 1.0 \\
&  & Relevance & 63.5 & 31.7 & 4.8 \\
\midrule
\multirow{3}{*}{T5 Large} & \multirow{3}{*}{T5 Base} & Personalization & 55.7 & 38.3 & 6.0 \\
&  & Quality & 56.8 & 40.9 & 2.3 \\
&  & Relevance & 52.9 & 41.0 & 6.1 \\
\hline
\end{tabular}}
\label{tab:amazon_book_win_rate}
	\vspace{-0.0in}
\end{table}

\noindent\textbf{Head-to-head records between T5 checkpoints.} Tables \ref{tab:amazon_book_win_rate} 
presents the head-to-head contest records between different T5 models as well as the human-written texts (GOLD) in all three evaluation dimensions on one of the datasets (more results are included in Appendix~\ref{sec:appendix-pairwise} and ~\ref{sec:appendix-finegrain}). We see that the human gold standard still has a higher (but not dominating) win rate against the best T5 generator in comparison. We see that when a generator model competes with another model smaller in size (and with a lower Elo rating), it has a higher winning rate. When the gap of Elo rating is larger, the winning rate increases. The LLM evaluator adeptly captures the degree of improvement a larger generator model brings compared to its smaller counterpart across all three dimensions. 

\subsection{Ablation Study}
Either at the test case level or at the generator level, {\sysname}'s evaluations on the three dimensions, quality, relevance, and personalization are more or less correlated. This is intuitive, as a more capable generator is likely to be better at all these aspects: writing fluently, keeping to the point, and mimicking an author. However, are these dimensions indeed measuring the relevance and personalization, or are they just a variant of the general quality? To validate that the three dimensions of {\sysname} are measuring what they are supposed to measure, we conduct controlled experiments and test whether {\sysname} scores are influenced by the nuanced differences between quality, relevance, and personalization (as defined in \ref{sec:multifacet}).


\begin{table}[!h]
\vspace{-0.1in}
	\caption{Ablation study by swapping user's personal context and swapping immediate context (Title) in generation. Original generator vs. ablated generators. Swapping personal context hurts personalization.  Swapping immediate context destroys relevance and reduces personalization.}	\vspace{-0.1in}
	\begin{center}
\resizebox{1\linewidth}{!}{\begin{tabular}{l|p{1cm}|p{1.1cm}|cp{0.5cm}||p{1cm}|p{1cm}|p{0.8cm}}
 	\toprule & \multicolumn{3}{c}{\textbf{Personal Context Swapped}} &&  \multicolumn{3}{c}{\textbf{Immediate Context Swapped}} \\

		\midrule \bf Eval Dim. & Win & Loss &Tie&& Win & Loss &Tie\\ 
\midrule
Personalization	&68.3&26.8&5.0&& 68.2&27.8&4.0\\
Quality&53.6&44.6&1.8&&51.8&46.5&1.7\\
Relevance&56.2&41.3&2.5&&96.4&3.3&0.3\\
		\midrule
\end{tabular}}
	\end{center}
\label{tab:ablation_user_history}
\vspace{-0.1in}
\end{table}


\noindent\textbf{Ablation study on personal context.} As personalization is evaluated by {\sysname}$_\texttt{Pers}$($\hat{X}^a_t$, $\hat{X}^b_t$ | $U_t$), a text example generated under a different user context $U_{t'}$ should not be as personalized (for the target user $t$) as the example generated under the target user's own context $U_t$, when all other conditions are identical. To verify this, we construct an ablated generator that uses the same fine-tuned T5 XXL model and test cases except for randomly swapping the user's historical writing examples with those of another user's. Table~\ref{tab:ablation_user_history} presents ablation results of how manipulating the user's writing history impacts generation quality. We see that the original generator has a close to 50-50 win rate against the ablated generator on quality and relevance, which is reasonable as everything except for $U_t$ is controlled. The ablated generator does slightly worse on these two dimensions, possibly because the T5 generator gets confused by the mismatch between the immediate context $Q$ (which is also composed by the user $t$) and the swapped user context $U_{t'}$.  As expected, the original model obtains a clearly higher win rate (68.3\%) in personalization over the ablated generator,  ranging from 57-83\% on different datasets (see Appendix~\ref{sec:appendix-ablation}). This indicates that the personal context provided to the T5 generator has a significant influence on how much the output is personalized, and this influence is captured by {\sysname}'s personalization evaluation. 

\noindent \textbf{Ablation study on immediate context.} Similarly, as relevance is evaluated by {\sysname}$_\texttt{Rel}$($\hat{X}^a_t$, $\hat{X}^b_t$ | $Q$), a text example generated given a different immediate context $Q'$ should not be as relevant to the original query $Q$, when all other conditions are identical. To verify this, we ablate the T5 XXL generator by randomly swapping the immediate context $Q$ in a test case with one from another test case, while all other conditions are controlled. Table~\ref{tab:ablation_context} presents the win-loss ratio of the original generator against the ablated generator. 
We see that while the win rate on quality is close to 50-50, the original generator has a dominating winning record (96.4\%) on relevance, as we expected.  
Interestingly, we also see inflated a win rate on personalization, although not as significant as those on relevance. This is because the immediate contexts in our case are the titles or starting sentences of the reviews or emails, which also encodes the author's writing styles. Swapping it with that of another author could also make the generated text less personalized. Nevertheless, the personal context $U_t$ still contains rich information, and the ablated generator is still able to generate texts in the user's voice but completely irrelevant.  This result verifies that {\sysname}'s relevance and personalization evaluations do capture the nuances in these two aspects that are separated from the general text quality. 
Interested readers may refer to Appendix~\ref{sec:appendix-ablation} for more details about the ablation study.

\subsection{Emerging Ability of LLM Evaluators }


\begin{figure}[!h]
    \centering

    \begin{subfigure}[b]{0.43\textwidth}
        \centering
        \includegraphics[width=\textwidth]{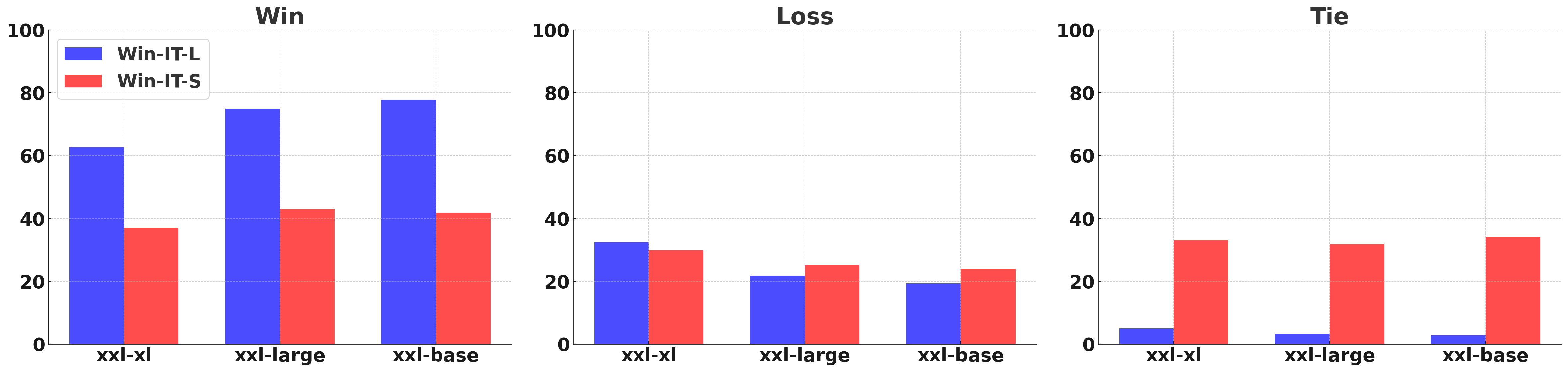} 
        \caption{Personalization}
        \label{fig:sub1}
    \end{subfigure}
    \hfill 
    \begin{subfigure}[b]{0.43\textwidth}
        \centering
       \includegraphics[width=\textwidth]{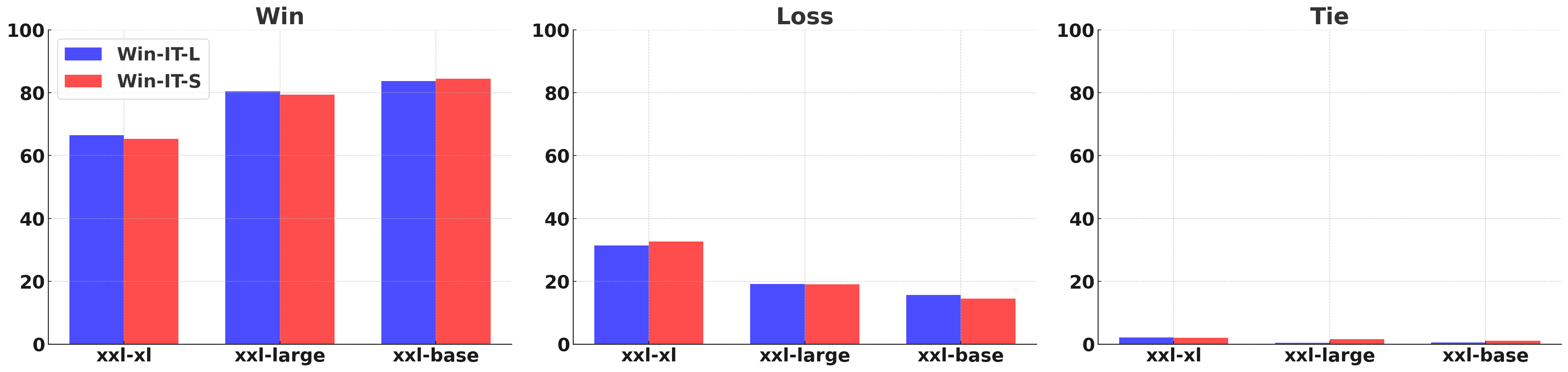}
        \caption{Quality}
        \label{fig:sub2}
    \end{subfigure}
    \hfill
    \begin{subfigure}[b]{0.43\textwidth}
        \centering
       \includegraphics[width=\textwidth]{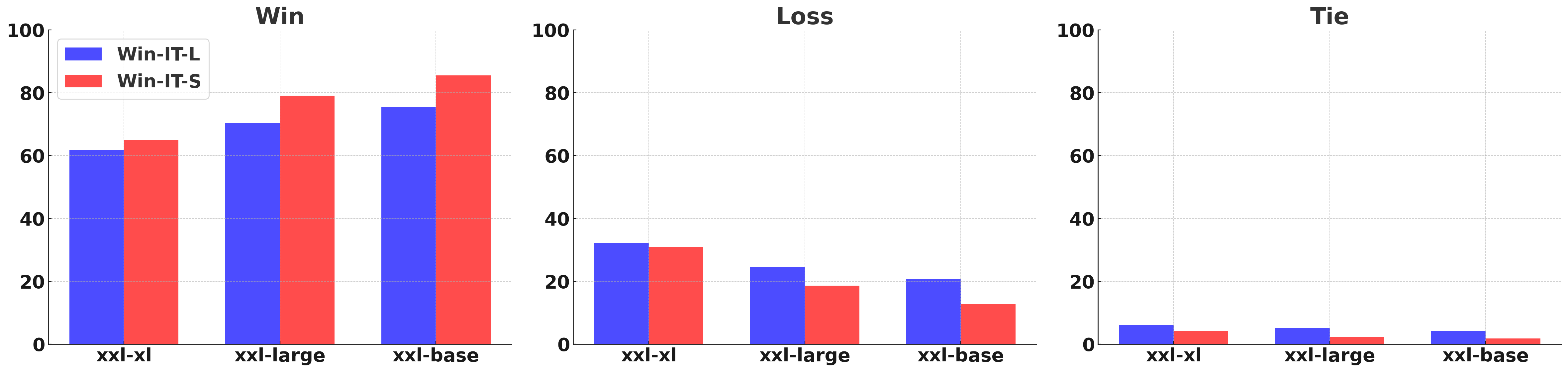} 
        \caption{Relevance}
        \label{fig:sub2}
    \end{subfigure}   
    \caption{Pairwise comparison of multiple T5 generators using PaLM 2-IT-L and PaLM 2-IT-S as evaluators. The Tie-rate in Personalization is significantly higher when switching the evaluator from PaLM 2-IT-L to PaLM 2-IT-S while Quality and Relevance evaluations are not significantly affected. }
   
    \label{fig:it_l_s}
    
\end{figure}

We also examine the use of different LLMs as the evaluator, in particular PaLM 2-IT-L versus PaLM 2-IT-S in PaLM 2 model family~\cite{anil2023palm} (with PaLM 2-IT-S being a smaller model), to compare the T5-XXL generator with three other generator models: T5-XL, T5-Large, and T5-Base. Figure~\ref{fig:it_l_s} shows the Win-Loss-Tie rates in personalziation, quality, and relevance for every matched pair. PaLM 2-IT-L and PaLM 2-IT-S produce similar evaluation results in terms of quality and relevance, showing the highest win rate when T5-XXL competes against Base, followed by a lower win rate against T5-Large and an even lower win rate against T5-XL (but still higher than 60\%). The two evaluators however produce very different result on personalization. PaLM 2-IT-L's judgments on personalization are similar to those on the other two dimensions, while PaLM 2-IT-S outputs much lower win rates in all three match-ups, as well as over 30\% of ties. 
This indicates that while the smaller evaluator PaLM 2-IT-S is as capable as the larger model to evaluate quality and relevance, it struggles to accurately assess personalziation, which is more subtle and subjective. As the size of the evaluator model grows, the ability of evaluating personalization emerges.

\vspace{0.05in}
\section{Conclusion and Discussion}
\vspace{0.07in}

We present the first automated framework that is specifically evaluating personalized text generation. The proposed framework, {\sysname}, formally distinguishes three pivotal aspects of the goodness of a generated text, namely the general \textit{quality}, \textit{relevance}, and the degree of \textit{personalization}, and it leverages large language models as evaluators to make preferential judgments on each of the three dimensions. The evaluations made by {\sysname} are more accurate, more consistent, and more sensitive than those of trained human raters as well as those obtained through traditional reference-based metrics used for natural language generation, and its decisions require a smaller sample of test cases and no human annotation. The Elo ratings based on {\sysname} provide an objective and robust metric to benchmark the progress of personalized text generation. 

While {\sysname} presents desirable advantages over human raters, it may not be taken as an evidence of ``superhuman performance of LLMs.'' This is because the human raters involved were not the target user of the personalization task, and they might not have the best knowledge to evaluate a content that is personalized for others. However, such as mismatch is common in NLG evaluation where the content being evaluated was usually not generated in the context of the human judges. {\sysname} fills this gap when it is infeasible or costly to recruit the original users as the judges. 

When the content generated by the target users are available, they can still be used as a gold-standard reference. However, traditional reference-based metrics such as BLEU and ROUGE are unable to distinguish the nuances between the different aspects of text quality beyond its lexical similarity to the reference. We recommend to use an (preferably larger) LLM evaluator to compare a model-generated content to the human-written reference. 

Our work focuses on evaluating text generation that is tailored for specific users. It is worth noting that alignment with personal facts is encapsulated within our concept of personalization evaluation. More generic factual assessments, such as hallucination checks, are critical issues of text generation but not the focus of this paper. 
Much prior work has delved into this dimension and contributed valuable insights (e.g., ~\cite{llm_eval,bohnet2022attributed}). Furthermore, while our framework emphasizes these three dimensions, quality, relevance, and personalization, they by no means cover all aspects of the goodness of generated text. The landscape of evaluating text generation is vast and continually evolving. Numerous other dimensions warrant exploration and could be incorporated into a similar evaluation framework like {\sysname} in future work.


\bibliographystyle{ACM-Reference-Format}
\bibliography{ref}

\appendix
\clearpage
\section{Appendix}

\subsection{Elo rating}
\label{sec:appendix-elo}
The essential idea of Elo rating is to update a player's rating \( R \) after every game they play according to the formula:
\[
R_{\text{new}} = R_{\text{old}} + K(S - E),
\]
where \( R_{\text{old}} \) is the player's previous rating and is initialized as $1,000$, \( K \) is the weight and is set to $4$ in our calculation, \( S \) is the actual outcome of the game (1 for a win, 0.5 for a draw, and 0 for a loss), \( E \) is the expected outcome of the match based on the formula:
\[
E = \frac{1}{1 + 10^{\left(\frac{R_{\text{opponent}} - R_{\text{old}}}{400}\right)}}
\]
where \( R_{\text{opponent}} \) is the rating of the opponent before the game. The Elo system dynamically adjusts a player's rating based on the outcomes of played games, incrementing the rating of the winner and decrementing that of the loser, with the amount of adjustment dependent on the difference in ratings between the players, thus ensuring that players have ratings that accurately reflect their skill levels over time. 

\subsection{Dataset}
\label{sec:appendix-data}
We follow the processing steps of LaMP~\cite{salemi2023lamp} to organize emails by sender addresses, which exceeds $279$ accounts as there may be external senders. For all the six datasets, we deduplicate identical documents from each user's personal context.  A document can be included in our document set for generation if it exceeds 300 characters in length and its author has previously generated at least 3 documents. We retain users with a minimum of 10 examples. To prevent datasets from being overly influenced by particularly active users, we limit the 100 examples per each user. In order to evaluate the model’s ability to generalize, we partition
the datasets by users so that the validation and test sets only contain documents from users that are unseen in the training set. The partition ratio of users in train/validation/test sets are 85/5/10.

\subsection{Prompts used in Generation}
\label{sec:appendix-prompt}
We have three distinct prompts for amazon review datasets, reddit and avocado email respectively. Table~\ref{table:dataset_prompts} includes the prompt context and $Q$ and $U_t$ are placeholder which are replaced by the concrete context of each examples. 




\begin{table*}[h]
\centering
\caption{Prompts used in Generation. $Q$ and $U_t$ indicate corresponding immediate context and user profile. }
\begin{tabular}{|p{0.25\linewidth}|p{0.6\linewidth}|}
\hline
\textbf{Dataset} & \textbf{Prompt} \\
\hline
\multirow{4}{*}{Amazon Review} & Using the style of the provided examples, compose a new Amazon review. Ensure that your review aligns with the given context. \\
& Context: \{$Q$\} \\
& User samples: \{$U_t$\} \\
\hline
\multirow{4}{*}{Reddit} & Using the style of the provided examples, compose a new Reddit post. Ensure that your post aligns with the given context. \\
& Context: \{$Q$\} \\
& User samples: \{$U_t$\} \\
\hline
\multirow{4}{*}{Avocado Email} & Using the style of the provided examples, compose a new email. Ensure that your email aligns with the given context. \\
& Context: \{$Q$\} \\
& User samples: \{U$_t$\} \\
\hline
\end{tabular}
\label{table:dataset_prompts}
\end{table*}

\subsection{Traditional NLG metrics across datasets}
\label{sec:appendix-metrics}
Table~\ref{tab:bleu_rouge} shows the performance of different models using BLEU and ROUGE metrics on several datasets. The performance ranking across different evaluation metrics, such as BLEU and the various ROUGE scores, is not consistent. For example, while the T5-XXL  model exhibits the highest BLEU score on the Avocado email dataset, it does not dominate in all ROUGE scores. Another intriguing observation is that model sizes, ranging from Base to XXL do not linearly correspond to its performance.  This deviation suggests the intricacies inherent in the evaluation of personalization. 
\begin{table}[ht]
\centering
\caption{BLEU and ROUGE results for various models on different datasets.}
	\vspace{-0.1in}
	\resizebox{1.0\linewidth}{!}{
\begin{tabular}{c|c|c|c|c|c}
\hline
\textbf{Dataset} & \textbf{Model Name} & \textbf{BLEU} & \textbf{ROUGE1} & \textbf{ROUGE2} & \textbf{ROUGEL} \\
\hline
\multirow{4}{*}{Avocado email} & xxl & 10.94 & 29.63 & 13.36 & 21.67 \\
                            & xl  & 10.27 & 29.70 & 13.01 & 21.42 \\
                            & large & 9.89 & 29.50 & 12.53 & 20.98 \\
                            & base & 10.33 & 27.90 & 13.08 & 21.37 \\
\hline
\multirow{4}{*}{Amazon books} & xxl & 4.82 & 29.04 & 6.84 & 17.69 \\
                              & xl  & 4.27 & 26.12 & 5.95 & 16.51 \\
                              & large & 3.67 & 23.64 & 5.06 & 15.76 \\
                              & base & 3.69 & 22.49 & 4.84 & 15.47 \\
\hline
\multirow{4}{*}{Amazon clothing} & xxl & 7.46 & 36.30 & 10.30 & 21.49 \\
                                 & xl  & 7.13 & 34.63 & 9.80 & 20.91 \\
                                 & large & 6.91 & 34.01 & 9.61 & 20.88 \\
                                 & base & 6.30 & 30.99 & 8.47 & 19.83 \\
\hline
\multirow{4}{*}{Amazon home} & xxl & 6.65 & 34.48 & 9.06 & 20.70 \\
                             & xl  & 6.35 & 32.13 & 8.51 & 19.96 \\
                             & large & 5.62 & 29.71 & 7.46 & 18.76 \\
                             & base & 5.51 & 28.18 & 7.19 & 18.72 \\
\hline
\multirow{4}{*}{Amazon movie} & xxl & 2.86 & 24.18 & 4.03 & 14.79 \\
                              & xl  & 2.47 & 20.23 & 3.26 & 13.47 \\
                              & large & 2.11 & 18.27 & 2.68 & 12.76 \\
                              & base & 1.84 & 15.94 & 2.25 & 11.97 \\
\hline
\multirow{4}{*}{Reddit} & xxl & 7.08 & 23.96 & 9.71 & 18.54 \\
                        & xl  & 7.75 & 21.58 & 9.34 & 17.86 \\
                        & large & 5.08 & 18.26 & 5.81 & 14.78 \\
                        & base & 7.50 & 19.65 & 8.72 & 17.20 \\
\hline
\end{tabular}}
\label{tab:bleu_rouge}
\end{table}

\subsection{Consistency and Sensitivity in Pairwise Model Comparison} 
\label{sec:appendix-consistency}
\begin{figure*}[!ht] 
\centering
\includegraphics[width=.33\textwidth]{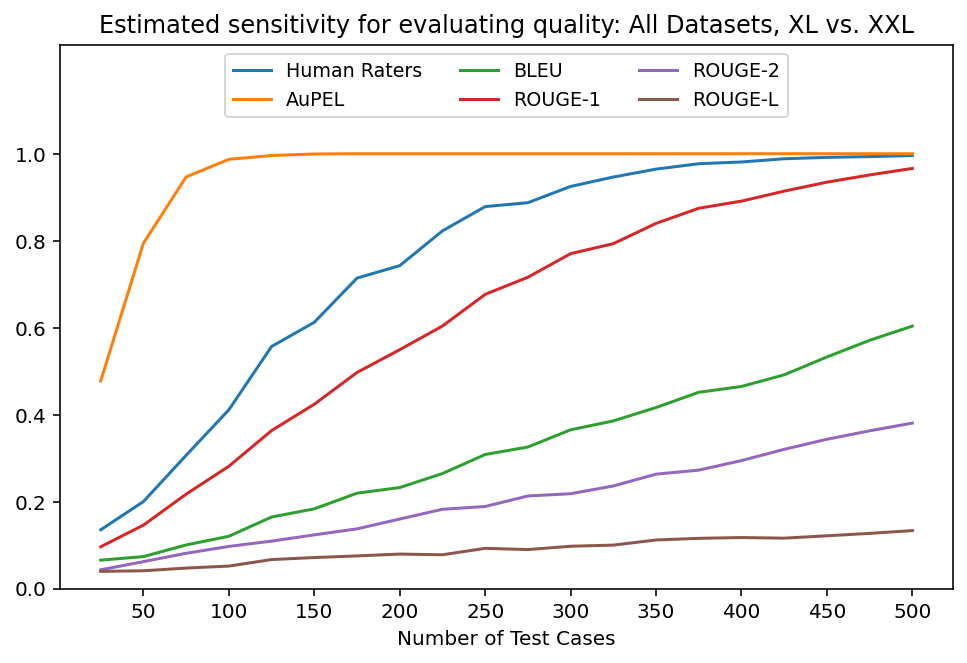}
\includegraphics[width=.33\textwidth]{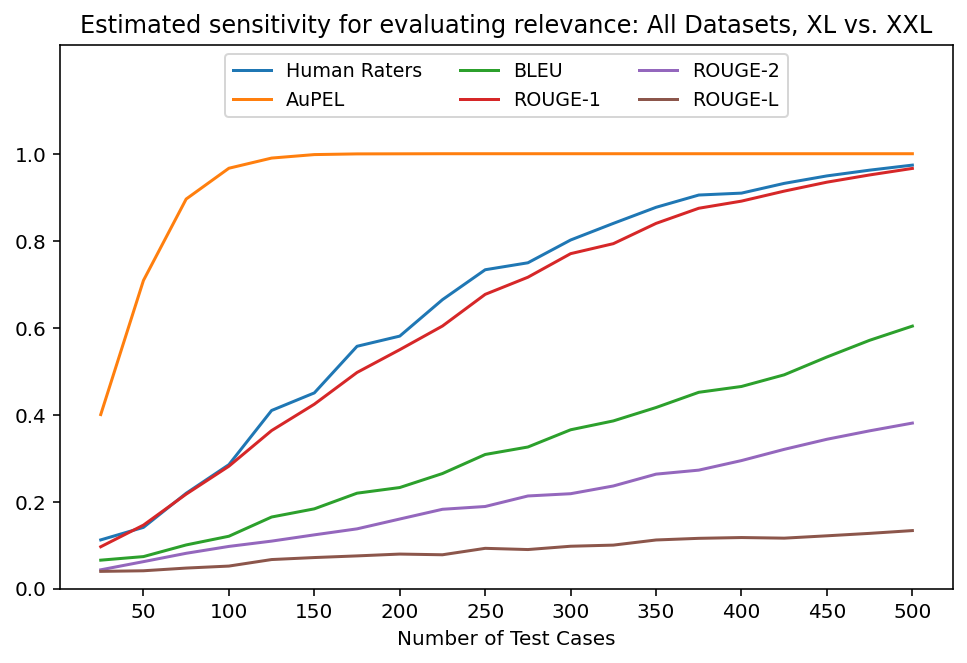}
\includegraphics[width=.33\textwidth]{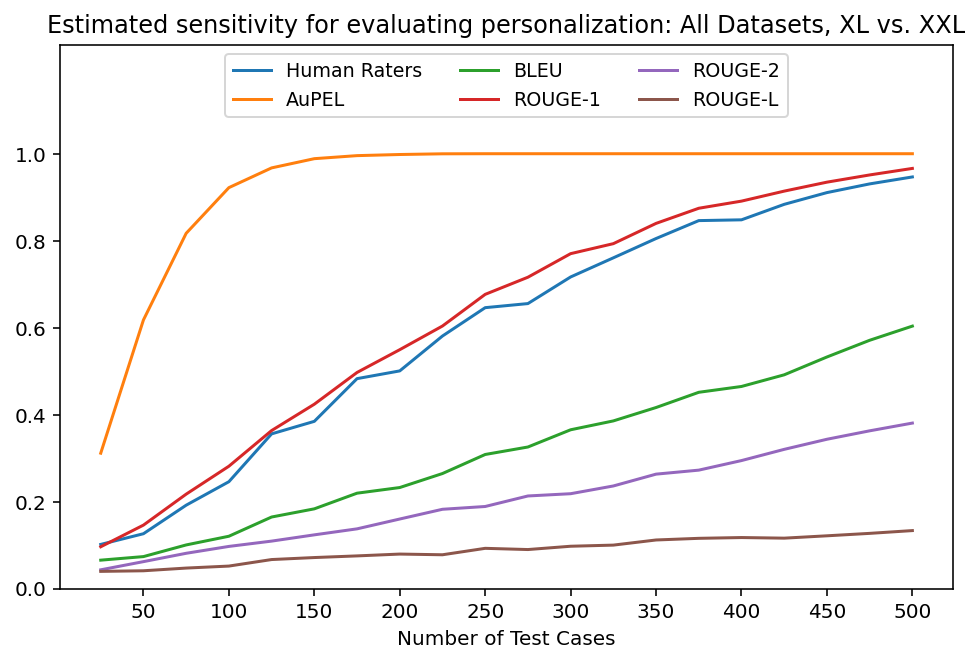}
\caption{Estimated sensitivity of {\sysname}, human raters, and NLG metrics for comparing T5 XL vs. T5 XXL generators.}
\label{figure:sensitivity_xl_xxl}
\end{figure*}

\begin{figure}[htbp]
	\vspace{-0.1in}
    \centering
    \begin{subfigure}[b]{0.28\textwidth}
        \centering
        \includegraphics[width=\textwidth]{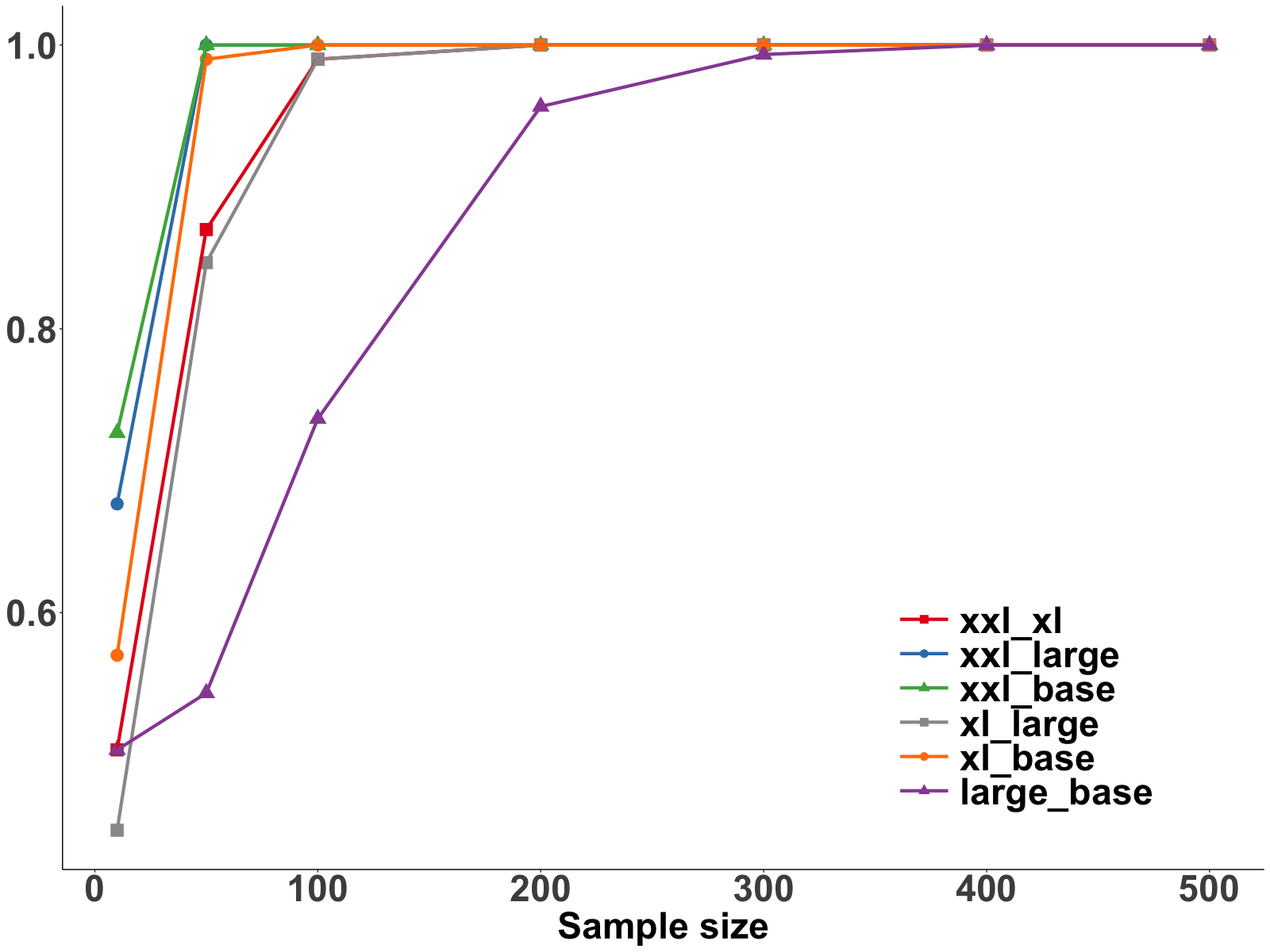} 
        \caption{Consistency}
        \label{fig:sub1}
    \end{subfigure}
    \hfill 
    \begin{subfigure}[b]{0.28\textwidth}
        \centering
       \includegraphics[width=\textwidth]{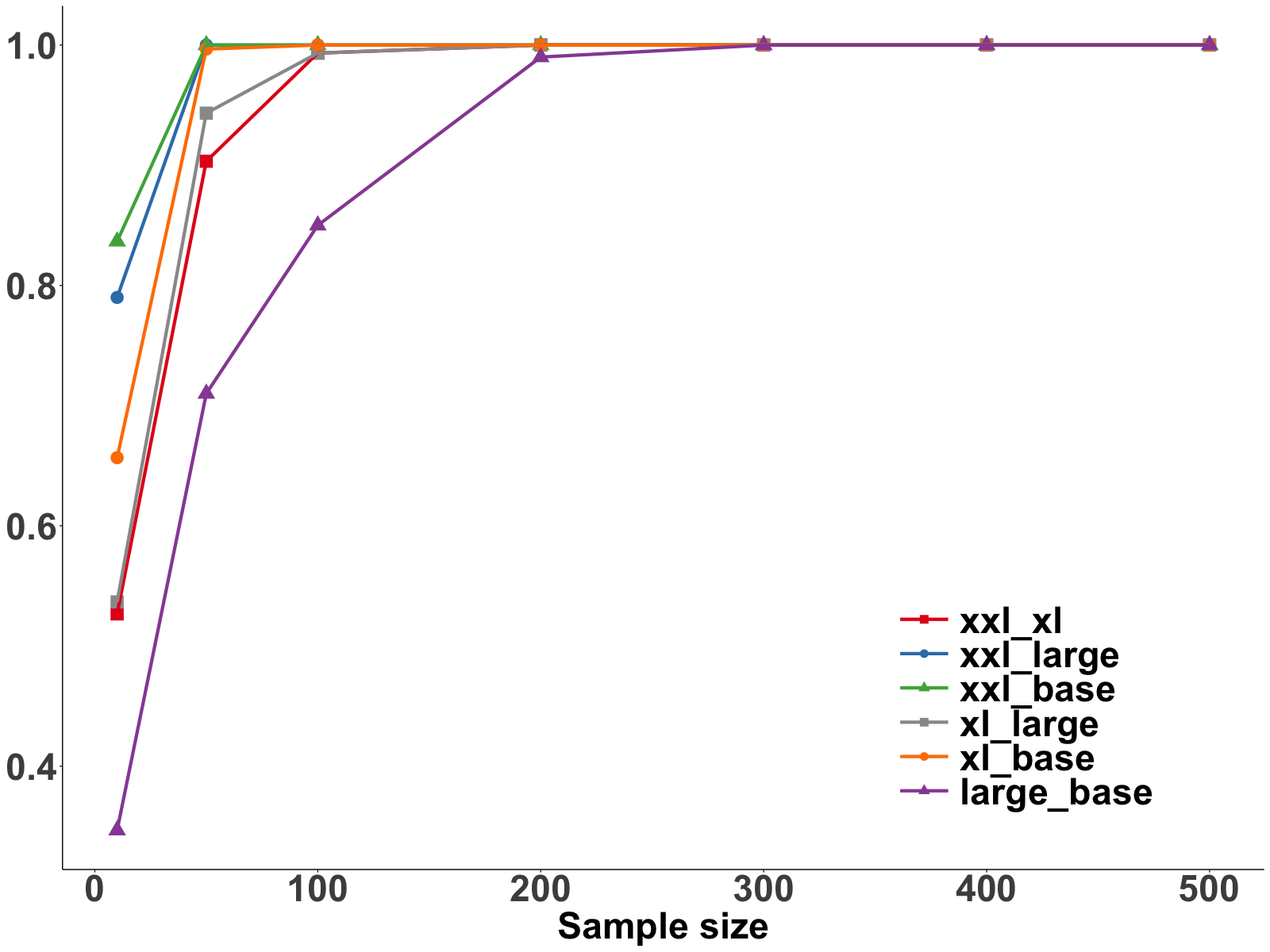} 
        \caption{Sensitivity}
        \label{fig:sub2}
    \end{subfigure}
    \caption{Consistency and Sensitivity Analysis of {\sysname} for various T5 checkpoints on Amazon book dataset.}
    \label{fig:consistency-sensitivity-all-pair}
    	\vspace{-0.1in}
\end{figure}

Our consistency (Figure~\ref{figure:consistency_xl_xxl}) and sensitivity (Figure~\ref{figure:sensitivity_xl_xxl}) analysis demonstrates strong evaluation properties across T5 model comparisons. We evaluate T5-XXL, T5-XL, T5-Large, and T5-Base and all model comparisons achieve 100\% consistency within hundreds of examples, highlighting the strong consistency and sensitivity of LLM evaluation. Further, we find the sample size required for discrimination scales with the capability gap between models (Figure~\ref{fig:consistency-sensitivity-all-pair}). Comparisons with wider gaps like T5-XXL versus T5-Base require fewer examples than closer models like T5-XL versus T5-Large. The ordering of required samples aligns with intuitive model strengths: XXL-Base > XXL-Large and XL-Base > XXL-XL and Large-Base. Overall, the consistency and sensitivity results align with model differences validate the reliability of LLM assessment. The discrimination scales predictably based on the mismatch between models, empirically confirming the methodology. Our analysis demonstrates efficient and stable quantification of model capabilities.

\subsection{Elo ratings with 95\% confidence interval}
\label{sec:appendix-elo-conf}
\begin{figure}[htbp] 
\centering   
\includegraphics[width=.3\textwidth]{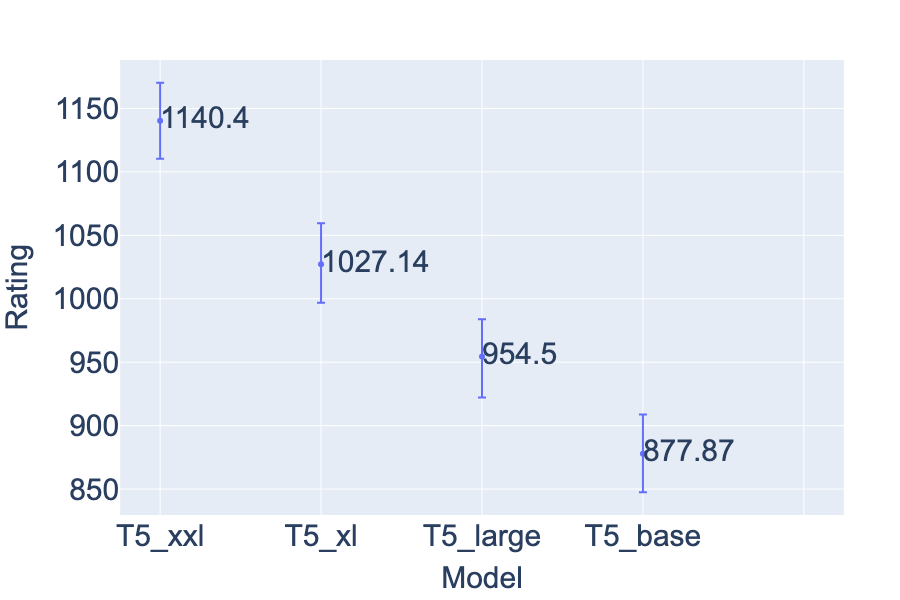}
\caption{Elo ratings of different checkpoints with 95\% confidence intervals.}  
	\vspace{-0.1in}
\label{fig:elo_errobound}
\end{figure}

Figure~\ref{fig:elo_errobound} shows the Elo ratings of various T5 generators with corresponding confidence intervals at 95\%. The performance rankings are as follows: T5 XXL, XL, Large, and Base checkpoints, in descending order from the largest to the smallest model. Notably, there is no overlap between the intervals for different checkpoints.

\subsection{Pairwise evaluation results between various T5 generators, by dataset} 
\label{sec:appendix-pairwise}
We evaluate T5 XXL, XL, Large and Base checkpoints in a pairwise manner and show the results in Tables~\ref{tab_appendix:book}, ~\ref{tab_appendix:clothing}, ~\ref{tab_appendix:home}, ~\ref{tab_appendix:moive}, ~\ref{tab_appendix:reddit} and ~\ref{tab_appendix:avocado_email}.

\subsection{Fine-grained Evaluation of Generators} 
\label{sec:appendix-finegrain}
An advantage of the multi-facet evaluation is the possibility to distinguish the abilities of a generator model on different aspects. We use PaLM 2-S-IT as our generation model and prompt it to generate personalized text without fine-tuning.  The results in Table~\ref{tab:amazon_book_win_rate_w_human} indicate that PaLM 2-IT-S obtains a high win rate (75.5\%) in \textit{quality} against the human written examples (GOLD generator). 
However, we can see that the \textit{relevance} and especially the degree of \textit{personalization} of the text generated by the LLM still trails human-written text by a large margin, showing that the LLM is better at generating fluent text than generating relevant and personalized content. Traditional NLG metrics fail to capture these nuanced differences, but multi-faceted evaluation provides fine-grained insights and guide the development of different writing abilities of the LLMs. 
The LLM evaluator is also able to provide detailed explanation of its judgment (we show examples in Table~\ref{tab:comparison}), which can further guide the generator to improve on the nuanced aspects associated with each dimension.  

\begin{table}[!h]
\centering
\caption{Pairwise comparison of PaLM 2-IT-S and Gold Standard (Human generator) on Amazon book dataset.}	
\resizebox{0.8\linewidth}{!}{
\begin{tabular}{c|c|c|c|c|c}
\hline
Model a & Model b & Eval Dim & Win & Loss & Tie \\
\hline
\multirow{3}{*}{PaLM 2-S-IT} &  & Personalization & 8.6 & 89.2 & 2.2 \\
& Human & Quality & 75.5 & 23.0 & 1.5 \\
&  & Relevance &28.3 & 68.7 & 3.0 \\
\hline
\end{tabular}}
\label{tab:amazon_book_win_rate_w_human}

\end{table}

\subsection{Ablation Studies, by dataset} 
\label{sec:appendix-ablation}

To validate that the three dimensions of {\sysname} are measuring what they are supposed to measure, we conduct controlled experiments and test whether {\sysname} scores are influenced by the nuanced differences between quality, relevance, and personalization. Table~\ref{tab:ablation_history} and Table~\ref{tab:ablation_context} shows the results of the controlled experiments by dataset, as a supplementary to Table~\ref{tab:ablation_user_history}. 

\begin{table}[!h]
	\caption{Ablation study by swapping user's personal context in generation, per dataset. Original generator vs. ablated generator. Swapping personal context hurts personalization. }	\vspace{-0.1in}
	\begin{center}
\resizebox{0.7\linewidth}{!}{\begin{tabular}{l|c|lc c cc}
		\toprule \bf Dataset & Eval Dim. & Win & Loss & Tie \\ 
		\midrule
		\multirow{3}{*}{\bf Amazon books} 
		& Personalization & 64.2 & 30.8 & 5.0 \\
		& Quality & 54.6 & 44.2 & 1.2 \\
		& Relevance & 54.8 & 43.1 & 2.1 \\ 
		\midrule
		\multirow{3}{*}{\bf Amazon clothing} 
		& Personalization & 70.3 & 22.3 & 7.4 \\
		& Quality & 49.2 & 49.0 & 1.8 \\
		& Relevance & 72.0 & 26.5 & 1.5 \\ 
		\midrule
		\multirow{3}{*}{\bf Amazon home} 
		& Personalization & 65.5 & 28.9 & 5.6 \\
		& Quality & 50.8 & 47.4 & 1.8 \\
		& Relevance & 50.5 & 46.7 & 2.8 \\ 
		\midrule
		\multirow{3}{*}{\bf Amazon movie} 
		& Personalization & 69.0 & 27.7 & 3.3 \\
		& Quality & 54.6 & 43.6 & 1.8 \\
		& Relevance & 54.0 & 44.8 & 1.2 \\ 
		\midrule
		\multirow{3}{*}{\bf Reddit} 
		& Personalization & 57.2 & 36.3 & 6.5 \\
		& Quality & 51.1 & 46.1 & 2.8 \\
		& Relevance & 54.0 & 42.3 & 3.7 \\ 
		\midrule
		\multirow{3}{*}{\bf Avocado email} 
		& Personalization & 83.4 & 14.7 & 1.9\\
		& Quality & 61.1 & 37.4 & 1.4 \\
		& Relevance & 51.7 & 44.5 & 3.8  \\ 
		\midrule
\end{tabular}}
	\end{center}
\label{tab:ablation_history}

\end{table}

\begin{table}[!h]
	\caption{Ablation study by swapping immediate context (Title and start of text) in generation, per dataset. Original generator vs. ablated generator. Swapping immediate context destroys relevance and moderately reduces personalization.}  
	\vspace{-0.1in}
	\begin{center}
	\resizebox{0.7\linewidth}{!}{
		\begin{tabular}{l|c|lc c cc}
			\toprule \bf Dataset & Eval Dim. &Win & Loss & Tie       \\ \midrule

		    \multirow{3}{*}{\bf Amazon books} &  Personalization & 68.3 & 28.7 & 3.0 \\
		& Quality & 54.7 & 44.5 & 0.8 \\
		& Relevance & 96.1 & 3.6 & 0.3 \\ 
		    \midrule
		      \multirow{3}{*}{\bf Amazon clothing} 	& Personalization & 61.1 & 34.0 & 4.9 \\
		& Quality & 47.1 & 50.3 & 2.6 \\
		& Relevance & 99.4 & 0.6 & 0.0 \\ 
		    \midrule
		       \multirow{3}{*}{\bf Amazon home} &  Personalization & 59.4 & 34.8 & 5.8 \\
		& Quality & 49.4 & 48.5 & 2.1 \\
		& Relevance & 99.9 & 0.1 & 0.0 \\ 
		    \midrule
		        \multirow{3}{*}{\bf Amazon movie} &   Personalization & 57.3 & 39.2 & 3.5 \\
		& Quality & 52.4 & 46.1 & 1.5 \\
		& Relevance & 87.7 & 11.2 & 1.1 \\ 
		 \midrule
		       \multirow{3}{*}{\bf Reddit} & Personalization & 92.8 & 6.2 & 1.0 \\
		& Quality & 51.5 & 47.2 & 1.3 \\
		& Relevance & 97.0 & 2.7 & 0.3 \\ 
		    \midrule
		       \multirow{3}{*}{\bf Avocado email} & Personalization & 70.6 & 23.7 & 5.7\\
		    & Quality& 55.9 & 42.2 & 1.9\\
		    & Relevance & 98.6 & 1.4 & 0.0 \\ 

			\bottomrule
		\end{tabular}}
	\end{center}
\label{tab:ablation_context}

\end{table}

\begin{table}[ht]
\centering
\caption{Pairwise evaluation results across various T5 check-points on Amazon book reivew dataset.}
\resizebox{0.6\linewidth}{!}{
\begin{tabular}{c|c|c|c|c|c}
\hline
Model a & Model b & Eval Dim & Win & Loss & Tie \\
\hline
\multirow{9}{*}{xxl} & xl & personalization & 62.6 & 32.4 & 5.0 \\
& xl & quality & 66.5 & 31.4 & 2.1 \\
& xl & relevance & 61.8 & 32.2 & 6.0 \\
\cline{2-6}
& large & personalization & 74.9 & 21.8 & 3.3 \\
& large & quality & 80.4 & 19.2 & 0.4 \\
& large & relevance & 70.4 & 24.5 & 5.1 \\
\cline{2-6}
& base & personalization & 77.8 & 19.4 & 2.8 \\
& base & quality & 83.7 & 15.7 & 0.6 \\
& base & relevance & 75.3 & 20.6 & 4.1 \\
\midrule
\multirow{6}{*}{xl} & large & personalization & 62.6 & 32.6 & 4.8 \\
& large & quality & 68.2 & 29.7 & 2.1 \\
& large & relevance & 59.5 & 34.1 & 6.4 \\
\cline{2-6}
& base & personalization & 68.3 & 27.5 & 4.2 \\
& base & quality & 73.4 & 25.6 & 1.0 \\
& base & relevance & 63.5 & 31.7 & 4.8 \\
\midrule
\multirow{3}{*}{large} & base & personalization & 55.7 & 38.3 & 6.0 \\
& base & quality & 56.8 & 40.9 & 2.3 \\
& base & relevance & 52.9 & 41.0 & 6.1 \\
\hline
\end{tabular}}
\label{tab_appendix:book}
\end{table}

\begin{table}[ht]
\centering
\caption{Pairwise evaluation results across various T5 check-points on Amazon clothing reivew dataset.}
\resizebox{0.6\linewidth}{!}{
\begin{tabular}{c|c|c|c|c|c}
\hline
Model a & Model b & Eval Dim & Win & Loss & Tie \\
\hline
\multirow{9}{*}{xxl} & xl & personalization & 59.9 & 33.8 & 6.3 \\
& xl & quality & 68.5 & 29.3 & 2.2 \\
& xl & relevance & 56.6 & 38.4 & 5.0 \\
\cline{2-6}
& large & personalization & 66.8 & 28.3 & 4.9 \\
& large & quality & 75.9 & 22.4 & 1.7 \\
& large & relevance & 59.6 & 35.1 & 5.3 \\
\cline{2-6}
& base & personalization & 74.4 & 20.4 & 5.2 \\
& base & quality & 88.8 & 9.8 & 1.4 \\
& base & relevance & 71.5 & 25.0 & 3.5 \\
\midrule
\multirow{6}{*}{xl} & large & personalization & 53.4 & 38.7 & 7.9 \\
& large & quality & 58.9 & 38.7 & 2.4 \\
& large & relevance & 50.6 & 43.1 & 6.3 \\
\cline{2-6}
& base & personalization & 67.8 & 27.3 & 4.9 \\
& base & quality & 79.3 & 19.2 & 1.5 \\
& base & relevance & 66.1 & 30.5 & 3.4 \\
\midrule
\multirow{3}{*}{large} & base & personalization & 62.4 & 32.1 & 5.5 \\
& base & quality & 75.0 & 23.6 & 1.4 \\
& base & relevance & 61.4 & 32.9 & 5.7 \\
\hline
\end{tabular}}
\label{tab_appendix:clothing}
\end{table}

\begin{table}[ht]
\centering
\caption{Pairwise evaluation results across various T5 check-points on Amazon home review dataset.}
\resizebox{0.6\linewidth}{!}{
\begin{tabular}{c|c|c|c|c|c}
\hline
Model a & Model b & Eval Dim & Win & Loss & Tie \\
\hline
\multirow{9}{*}{xxl} & xl & personalization & 63.4 & 29.1 & 7.5 \\
& xl & quality & 67.7 & 29.5 & 2.8 \\
& xl & relevance & 58.7 & 35.1 & 6.2 \\
\cline{2-6}
& large & personalization & 74.9 & 19.9 & 5.2 \\
& large & quality & 81.4 & 17.4 & 1.2 \\
& large & relevance & 69.4 & 26.3 & 4.3 \\
\cline{2-6}
& base & personalization & 82.3 & 14.2 & 3.5 \\
& base & quality & 88.1 & 11.0 & 0.9 \\
& base & relevance & 77.8 & 18.3 & 3.9 \\
\midrule
\multirow{6}{*}{xl} & large & personalization & 60.7 & 35.2 & 4.1 \\
& large & quality & 65.1 & 32.7 & 2.2 \\
& large & relevance & 58.1 & 36.3 & 5.6 \\
\cline{2-6}
& base & personalization & 69.6 & 26.1 & 4.3 \\
& base & quality & 76.6 & 22.4 & 1.0 \\
& base & relevance & 65.6 & 30.1 & 4.3 \\
\midrule
\multirow{3}{*}{large} & base & personalization & 59.5 & 35.3 & 5.2 \\
& base & quality & 62.3 & 35.4 & 2.3 \\
& base & relevance & 58.4 & 35.6 & 6.0 \\
\hline
\end{tabular}}
\label{tab_appendix:home}
\end{table}

\begin{table}[htb]
\centering
\caption{Pairwise evaluation results across various T5 check-points on Amazon movie review dataset.}
\resizebox{0.6\linewidth}{!}{
\begin{tabular}{c|c|c|c|c|c}
\hline
Model a & Model b & Eval Dim & Win & Loss & Tie \\
\hline
\multirow{9}{*}{xxl} & xl & personalization & 68.0 & 27.8 & 4.2 \\
& xl & quality & 71.9 & 26.8 & 1.3 \\
& xl & relevance & 69.0 & 26.5 & 4.5 \\
\cline{2-6}
& large & personalization & 76.4 & 20.3 & 3.3 \\
& large & quality & 82.4 & 16.5 & 1.1 \\
& large & relevance & 76.6 & 19.0 & 4.4 \\
\cline{2-6}
& base & personalization & 82.2 & 15.4 & 2.4 \\
& base & quality & 89.5 & 9.8 & 0.7 \\
& base & relevance & 81.9 & 14.5 & 3.6 \\
\midrule
\multirow{6}{*}{xl} & large & personalization & 59.8 & 34.9 & 5.3 \\
& large & quality & 62.8 & 34.6 & 2.6 \\
& large & relevance & 59.3 & 35.3 & 5.4 \\
\cline{2-6}
& base & personalization & 71.2 & 25.6 & 3.2 \\
& base & quality & 76.0 & 21.8 & 2.2 \\
& base & relevance & 67.5 & 25.0 & 7.5 \\
\midrule
\multirow{3}{*}{large} & base & personalization & 58.4 & 35.4 & 6.2 \\
& base & quality & 62.8 & 34.2 & 3.0 \\
& base & relevance & 57.7 & 35.1 & 7.2 \\
\hline
\end{tabular}}
\label{tab_appendix:moive}
\end{table}

\begin{table}[ht]
\centering
\caption{Pairwise evaluation results across various T5 check-points on reddit dataset.}
\resizebox{0.6\linewidth}{!}{
\begin{tabular}{c|c|c|c|c|c}
\hline
Model a & Model b & Eval Dim & Win & Loss & Tie \\
\hline
\multirow{9}{*}{xxl} & xl & personalization & 67.1 & 28.6 & 4.3 \\
& xl & quality & 71.7 & 26.9 & 1.4 \\
& xl & relevance & 65.4 & 29.2 & 5.4 \\
\cline{2-6}
& large & personalization & 80.4 & 17.9 & 1.7 \\
& large & quality & 80.4 & 18.2 & 1.4 \\
& large & relevance & 71.0 & 24.2 & 4.8 \\
\cline{2-6}
& base & personalization & 84.6 & 13.7 & 1.7 \\
& base & quality & 83.3 & 15.7 & 1.0 \\
& base & relevance & 74.4 & 21.6 & 4.0 \\
\midrule
\multirow{6}{*}{xl} & large & personalization & 60.2 & 36.1 & 3.7 \\
& large & quality & 59.7 & 37.7 & 2.6 \\
& large & relevance & 51.2 & 41.1 & 7.7 \\
\cline{2-6}
& base & personalization & 72.2 & 24.9 & 2.9 \\
& base & quality & 66.6 & 31.1 & 2.3 \\
& base & relevance & 59.9 & 33.8 & 6.3 \\

\midrule
\multirow{3}{*}{large} & base & personalization & 59.7 & 36.7 & 3.6 \\
& base & quality & 59.3 & 39.3 & 1.4 \\
& base & relevance & 53.2 & 38.6 & 8.2 \\
\hline
\end{tabular}}
\label{tab_appendix:reddit}
\end{table}

\begin{table}[ht]
\centering
\caption{Pairwise evaluation results across various T5 check-points on Avocado email dataset.}
\resizebox{0.6\linewidth}{!}{
\begin{tabular}{c|c|c|c|c|c}
\hline
Model a & Model b & Eval Dim & Win & Loss & Tie \\
\hline
\multirow{9}{*}{xxl} & xl & personalization & 52.13 & 41.71 & 6.16 \\
& xl & quality & 61.14 & 36.97 & 1.90 \\
& xl & relevance & 54.50 & 42.65 & 2.84 \\
\cline{2-6}
& large & personalization & 55.45 & 41.23 & 3.32 \\
& large & quality & 64.93 & 34.12 & 0.95 \\
& large & relevance & 55.92 & 42.65 & 1.42 \\
\cline{2-6}
& base & personalization & 64.93 & 32.70 & 2.37 \\
& base & quality & 73.93 & 24.64 & 1.42 \\
& base & relevance & 63.51 & 33.65 & 2.84 \\
\midrule
\multirow{6}{*}{xl} & large & personalization & 45.97 & 47.39 & 6.64 \\
& large & quality & 55.45 & 41.23 & 3.32 \\
& large & relevance & 48.82 & 47.87 & 3.32 \\
\cline{2-6}
& base & personalization & 59.24 & 36.02 & 4.74 \\
& base & quality & 63.51 & 33.18 & 3.32 \\
& base & relevance & 59.24 & 39.34 & 1.42 \\
\midrule
\multirow{3}{*}{large} & base & personalization & 55.92 & 36.02 & 8.06 \\
& base & quality & 56.40 & 41.23 & 2.37 \\
& base & relevance & 57.35 & 40.28 & 2.37 \\
\hline
\end{tabular}}
\label{tab_appendix:avocado_email}
\end{table}

\clearpage

\begin{table*}[ht]
\centering
\caption{Evaluation Output of Sample 1 and Sample 2 via {\sysname} on the personalization dimension. The Judgments and Explanations 1 to 4 are generated by different runs of the LLM evaluator.}

\begin{tabular}{|p{0.25\linewidth}|p{0.6\linewidth}|}
\hline
\textbf{User profile} & improve that marathon pr!...5 stars for the actual book vs 4 stars for the kindle version \   my husband recently bought the kindle version of this book with a goal of improving his race time.  his upcoming late winter marathon will be his 3rd in about 7-8 months, so pete and scott's multiple-marathon-recovery-program was immediately put to use.  he has just finished the 3-week recovery program and is just now starting the 18-week/55-mile program to prepare for his next race.  the book presents an overview of all the thoughts and science behind these training programs. even as someone who keeps abreast of the latest on running, my husband felt that he gained a new understanding and dispelled some myths about the physiology of running and the right way to train.  when you get further into the book, you'll find the various", 
\\
\midrule
\textbf{Immeidate Context} & pretty good "jumping off" point for color mixing. color mixing recipes for landscapes: mixing recipes for more than 500 color combinations follow these 4 simple steps to mix more than 500 color combinations: step one look in the color guidance index for the subject you want to paintfor example, lilac, purplish. step two find the color recipe with the subjects recipe number (268) and a photo of the actual paint mixture \\
\midrule
\textbf{Sample A} &
i've been using this book for a while now and find it to be a pretty good ``jumping off'' point for mixing colors. i've used it to mix a number of the ``hard to get'' colors for my landscape paintings...and it's been helpful. i've also used it to mix some of the ``impossible'' colors...those colors that are so close to the original that you'd never think of mixing them yourself. \\
\midrule
\textbf{Sample B} & i have 2 of william powell's other color-mixing books.  i have found his color mixing recipes for portraits to be very useful; but felt his generic color mixing recipes book was a waste of money and as a result had initially hesitated to order any more of his books.  however, i was pleased when his landscape book arrived.  it gives a little bit of \"how-to\" info at the beginning of the book; but its main value is in all the numerous index listings in the back of the book and their correlated color \"recipes\" in the main part of the book.  for example, the index lists dozens of options for sky at various times of the day (early morning, sunset, low sun, twilight, you name it) and weather conditions with suggestions for recipes for clouds, sunlight, moon, etc.  for snow, it offers several variations based upon the amount of light hitting the snow (time of day) and provides recipes for main color, shadowing, and highlights.  your final mixed colors following these recipes are likely to vary \\
\hline
\textbf{Explanation 1} & Response (B) is likely to be written by the same author with given input example: both are long and detailed. Response (A) is short and bland.\\
\midrule
\textbf{Explanation 2} & Response (B) talks about a book and its value like the input example, while response (A) doesn't\\
\midrule
\textbf{Explanation 3} & Response (B) is likely to be written by the same author with given input example. Both are lengthy, have a lot of details, and include the authors' thoughts.\\
\midrule
\textbf{Explanation 4} & Both response (B) and the input example are long and detailed. They also have a similar writing style. For example, both use a lot of parentheses and dashes.\\
\midrule
\end{tabular}
\label{tab:comparison}
\end{table*}

\end{document}